\documentclass{article}

\usepackage{microtype}
\usepackage{graphicx}
\usepackage{subfigure}
\usepackage{booktabs} 

\usepackage{hyperref}

\usepackage[accepted]{icml2025}

\usepackage{amsmath}
\usepackage{amssymb}
\usepackage{mathtools}
\usepackage{amsthm}

\usepackage{multirow}
\usepackage{multicol}
\usepackage{xcolor}
\usepackage{colortbl}
\usepackage{pifont}
\usepackage{float}
\usepackage{algorithm}
\usepackage{algorithmicx}
\usepackage{algpseudocode}
\usepackage{caption}
\usepackage{subcaption}

\newcommand{\tcj}{\textcolor{black}}
\newcommand{\red}[1]{\textcolor{red}{#1}}

\usepackage[capitalize,noabbrev]{cleveref}

\theoremstyle{plain}
\newtheorem{theorem}{Theorem}[section]
\newtheorem{proposition}[theorem]{Proposition}
\newtheorem{lemma}[theorem]{Lemma}
\newtheorem{corollary}[theorem]{Corollary}
\theoremstyle{definition}
\newtheorem{definition}[theorem]{Definition}
\newtheorem{assumption}[theorem]{Assumption}
\theoremstyle{remark}
\newtheorem{remark}[theorem]{Remark}

\usepackage[textsize=tiny]{todonotes}

\icmltitlerunning{Attention Reallocation: Towards Zero-cost and Controllable Hallucination Mitigation of MLLMs}

\begin{document}
\twocolumn[
\icmltitle{Attention Reallocation: Towards Zero-cost and Controllable \\ Hallucination Mitigation of MLLMs}




\icmlsetsymbol{equal}{*}
\icmlsetsymbol{corresponding}{$\dagger$}
\begin{icmlauthorlist}
\icmlauthor{Chongjun Tu}{sch1,equal}
\icmlauthor{Peng Ye}{sch2,comp1,equal}
\icmlauthor{Dongzhan Zhou}{comp1}
\icmlauthor{Lei Bai}{comp1}
\icmlauthor{Gang Yu}{comp2}
\icmlauthor{Tao Chen}{sch1,corresponding}
\icmlauthor{Wanli Ouyang}{sch2,comp1}\\
$^{1}$ Fudan University \quad $^{2}$ The Chinese University of Hong Kong \quad $^{3}$ Shanghai Artificial Intelligence Laboratory \quad $^{4}$ StepFun
\end{icmlauthorlist}

\icmlcorrespondingauthor{Tao Chen}{eetchen@fudan.edu.cn}

\vskip 0.3in
]



\printAffiliationsAndNotice{\icmlEqualContribution} 

\begin{abstract}
Multi-Modal Large Language Models (MLLMs) stand out in various tasks but still struggle with hallucinations. 
While recent training-free mitigation methods mostly introduce additional inference overhead via retrospection strategy and contrastive decoding, we propose attention reallocation (AttnReal) to mitigate hallucinations with nearly zero extra cost.
Our approach is motivated by the key observations that, MLLM's unreasonable attention distribution causes features to be dominated by historical output tokens, which further contributes to hallucinated responses because of the distribution gap between different token types.
Based on the observations, AttnReal recycles excessive attention from output tokens and reallocates it to visual tokens, which reduces MLLM's reliance on language priors and ensures the decoding process depends more on the visual inputs.
More interestingly, we find that, by controlling the intensity of AttnReal, we can achieve a wide-range trade-off between the response faithfulness and overall performance.
Comprehensive results from different benchmarks validate the effectiveness of AttnReal across six open-source MLLMs and three decoding strategies.


\end{abstract}

\begin{figure*}[htb]
    \centering
    \includegraphics[width=\textwidth]{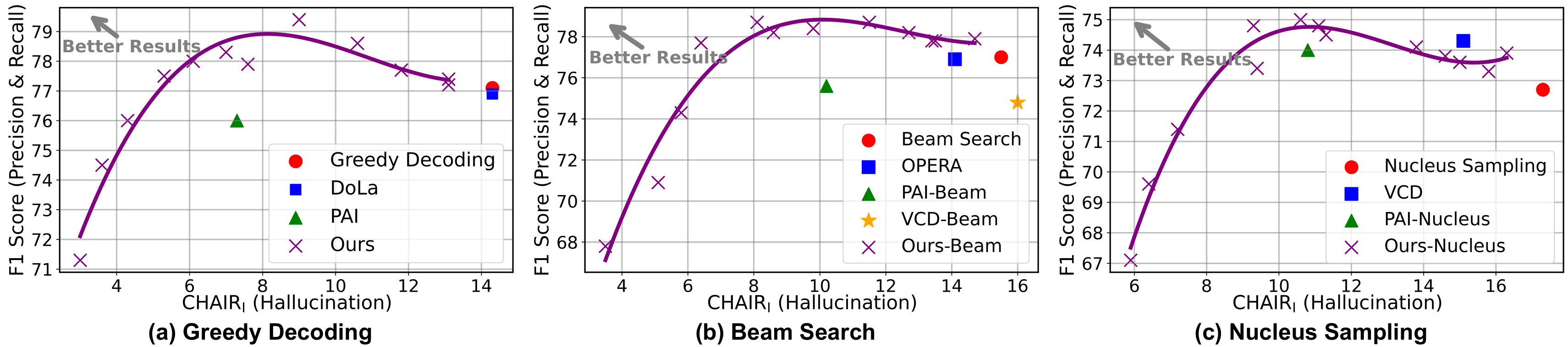}
    \vspace{-7mm}
    \caption{
    Performance comparison for various training-free methods to mitigate MLLM hallucinations on LLaVA-1.5-7B and the CHAIR benchmark using three decoding strategies.
    Lower CHAIR$_\text{I}$ represents fewer hallucinations. Higher F1 scores mean better \tcj{overall performance} (precision and recall).
    The curve in each sub-figure is obtained by adjusting the intensity of our method.
    Our proposed method not only yields superior results than state-of-the-art methods on different decoding strategies, but also achieves an excellent trade-off between hallucination and F1.
    }
    \label{fig: figure1}
    \vspace{-4mm}
\end{figure*}

\vspace{-3mm}
\section{Introduction}
\label{sec:intro}

Recently, Multimodal Large Language Models (MLLMs) have achieved remarkable progress, demonstrating superior performance in various tasks like image captioning, visual question answering, and complex content generation~\cite{han2023chartllama,brooks2023instructpix2pix,li2024llavamed,li2024clipsampowerful,cao2024madtppowerful,yang2024moosepowerful}.
However, MLLMs encounter a critical challenge known as hallucination, where their responses deviate from the input information, such as describing non-existent objects.
Such inaccuracies pose critical risks in safety-sensitive scenarios like autonomous driving~\cite{chen2024driving,cui2024driving,shao2024lmdriving} and AI-assisted healthcare~\cite{he2023health,wang2023chatcadhealth,li2024llavamed}, undermining MLLMs' reliability.

One effective approach to mitigating hallucinations in MLLMs is fine-tuning the models with carefully curated datasets~\cite{jiang2024hacltrain, wang2024mitigatingtrain,yin2024lammdataset,li2024enhanceddataset}, which is usually labor-intensive and requires substantial computational resources.
Alternatively, training-free methods are proposed to eliminate the need for additional data.
Early methods introduce new decoding strategies like Beam Search~\cite{graves2012sequencebeam,sutskever2014sequencebeam} and Nucleus Sampling~\cite{holtzman2019curiousnucleus} to increase the likelihood of faithful responses. 
Recent training-free methods primarily modify the decoding process, applying penalty terms \tcj{in the beam scores}~\cite{huang2024opera} or using contrastive decoding~\cite{leng2024vcd, liu2024pai, chuang2023dola,liang2024mitigating} to reduce MLLMs' over-reliance on statistical biases and language priors, which are regarded as one of the major hallucination origins~\cite{zhou2023analyzingreliance,leng2024vcd,zhai2023hallereliance,guan2024hallusionbenchreliance}.
However, these methods require careful design of contrast objects and introduce extra decoding steps, increasing latency and memory overhead.
Under such a background, a question naturally arises:

\textit{\textbf{Is it possible to mitigate MLLM hallucinations and improve the performance with nearly zero extra cost, without utilizing additional data or increasing decoding steps?}}

To answer this question, we investigate the origin of MLLM hallucinations in depth from the perspective of self-attention layers and different token types.
We obtain two key observations as follows, which are explained in detail in~\cref{sec:observations}.
\begin{itemize}
    \item 
    MLLM's unreasonable attention allocation between different token types (the input sequence consists of system, visual, instruction, and historical output tokens) 
    gradually amplifies the reliance on language priors, resulting in output tokens dominated features.
    \item 
    There is a significant gap between the feature distribution of visual tokens and textual tokens, so predicting the next token with features that over-rely on output tokens rather than input visual evidence tends to yield hallucinated responses.
\end{itemize}

Inspired by our findings, we propose a novel attention reallocation (AttnReal) approach, which mitigates hallucinations by redistributing attention values across different token types.
Specifically, AttnReal (1) detects and suppresses the attention sinks, which refer to tokens receiving significantly higher attention scores than others, in historical output tokens; (2) reallocates their reduced attention uniformly to visual tokens.
AttnReal is plug-and-play and simple-but-effective, demanding no extra data and almost no additional inference or memory overhead.
With AttnReal, we suppress the dominance of attention to historical output tokens, ensure decoded features in the forward process are better grounded in visual evidence, and improve the predicted logits, thereby mitigating the hallucinations.
Besides, since AttnReal requires only modifications to the attention layers, it can be widely applied to various MLLMs and different decoding strategies.
More interestingly, by adjusting the reallocation intensity from the attention sinks to visual tokens, a trade-off between response faithfulness and overall performance can be easily achieved.

Comprehensive results across hallucination benchmarks and GPT-assisted evaluations confirm AttnReal’s effectiveness.
As shown in~\cref{fig: figure1}, AttnReal achieves a controllable performance trade-off, yielding superior results to the state-of-the-art methods for both faithfulness and F1 scores across a relatively wide range of AttnReal intensities.
Moreover,~\cref{tab: decoding overhead} compares the decoding overheads of different methods, demonstrating the efficiency advantage of our proposed method.
Our contributions can be summarized as follows:
\begin{itemize}
    \item \textbf{The devil of MLLMs' hallucinations lies in the attention layers.}
    We investigate the attention and feature distribution from the perspective of self-attention layers, and establish the correlation between MLLM's unreasonable attention allocation and hallucinations.    
    \item \textbf{Attention reallocation as a free lunch to reduce MLLM hallucinations.}
    AttnReal achieves efficient and effective hallucination mitigation by reallocating attention from output sinks to visual tokens. It demands nearly zero extra overhead.
    \item \textbf{Plug-and-play capability and wide applicability.}
    AttnReal is applied to six open-source MLLMs and three decoding strategies, demonstrating excellent hallucination mitigation performances while improving overall performance under wide ranges of AttnReal intensities.   
\end{itemize}

\begin{table}[tbp]
\caption{Decoding overhead comparison among different methods. Greedy refers to the baseline Greedy Decoding. OPERA tracks multiple candidates at every decoding step. VCD, DoLa, and PAI involve contrastive decoding, which introduces additional decoding steps. As a comparison, our proposed method demands minimal extra overhead.}
\label{tab: decoding overhead}
\vspace{-2mm}
\resizebox{\columnwidth}{!}{%
\setlength{\tabcolsep}{2mm}
\begin{tabular}{lccc}
\toprule
Method &
  \begin{tabular}[c]{@{}c@{}}Contrastive \\ Decoding\end{tabular} &
  \begin{tabular}[c]{@{}c@{}}Candidates \\ per Step\end{tabular} &
  \begin{tabular}[c]{@{}c@{}}Decoding Overhead \\ (GFLOPs) \end{tabular} \\ 
\cmidrule(lr){1-4}
Greedy          & \ding{55} & 1 & 102 / token \\
OPERA~\cite{huang2024opera}           & \ding{55} & 5 & 1036 / token \\
VCD~\cite{leng2024vcd}             & \ding{51} & 1 & 214 / token \\
DoLa~\cite{chuang2023dola}            & \ding{51} & 1 & 135 / token \\
PAI~\cite{liu2024pai}             & \ding{51} & 1 & 126 / token \\
\textbf{Ours}   & \textbf{\ding{55}} & \textbf{1} & \textbf{104 / token} \\ 
\bottomrule
\end{tabular}%
}
\vspace{-4mm}
\end{table}

\section{Related Work}
\label{sec:related}

\subsection{Multi-Modal Large Language Models}
Recent advancements in multi-modal large-scale foundation models have greatly improved their cross-modal understanding and generation abilities. 
Early vision-language models based on BERT~\cite{kenton2019bert} combine visual and textual data to learn image-text interactions. 
The rise of open-source LLMs, like LLaMA~\cite{touvron2023llama,touvron2023llama2} and Vicuna~\cite{chiang2023vicuna}, contributes to the development of MLLMs such as LLaVA~\cite{liu2024visualllava} and MiniGPT-4~\cite{zhu2023minigpt} that process and generate multi-modal content through instruction tuning.
Recent advancements focus on specialized capability enhancements, including spatial grounding (Shikra~\cite{chen2023shikra}), modular modality collaboration (mPLUG-Owl2~\cite{ye2024mplug}), and cross-modal position embedding (Qwen2-VL~\cite{wang2024qwen2}). 
Despite these advancements, hallucination remains a persistent challenge across MLLMs. In this paper, we validate the effectiveness of our proposed method on these models.

\subsection{Hallucinations in MLLMs}
\noindent \textbf{Hallucination and the assessment.}
MLLM hallucinations refer to generated contents irrelevant or inconsistent with visual inputs, which undermines the models' reliability and practicality~\cite{yin2023woodpecker,zhou2023lure}.
Benchmarks like CHAIR~\cite{rohrbach2018chair}, MMHal-Bench~\cite{sun2023aligning_train_mmhal}, and AMBER~\cite{wang2023amber} assess hallucinations by comparing the generated content to the visual inputs and consider metrics like recall and precision to further measure the overall \tcj{performance}. 
Employing advanced GPT models to assist in evaluations is also a common practice~\cite{huang2024opera,liu2024pai}.

\noindent \textbf{Mitigation solutions.} 
Efforts to mitigate MLLM hallucinations can be categorized as training-based and training-free methods. 
Training-based methods fine-tune models on curated datasets~\cite{liu2023mitigatingtrain,gunjal2024detectingtrain} or introduce human feedback~\cite{yu2024rlhf}, but usually demand significant resources for data annotation and computation.
Early training-free methods introduce different decoding strategies, including Beam Search~\cite{boulanger2013audiobeam}, which tracks multiple candidate sequences, and Nucleus Sampling~\cite{holtzman2019curiousnucleus} that emphasizes coherent and varied outputs.
To further mitigate hallucinations, which are found to be associated with MLLM's over-reliance on statistical bias and language priors~\cite{yue2024lessismore}, various improvements have been proposed recently.
OPERA~\cite{huang2024opera} improves Beam Search with a penalty term \tcj{in the beam scores as well as a retrospection strategy}.
Other studies apply contrastive decoding to calibrate the MLLM's predicted logits:
DoLa~\cite{chuang2023dola} compares layer-wise predictions, HALC~\cite{chen2024halc} compares predictions between visual context windows, 
VCD~\cite{leng2024vcd} and PAI~\cite{liu2024pai} construct contrastive inputs by distorting and removing the input images, respectively.
While training-free, these methods incur inference overhead from tracking several candidates or multiple decoding steps.
In contrast, our proposed AttnReal suppresses hallucinations within a single forward pass, eliminating additional memory overhead and inference latency. Moreover, our method allows for a wide-range controllable trade-off between response faithfulness and \tcj{overall performance}, catering to different application scenarios.

\subsection{Attention Mechanism about MLLM Hallucinations}
\tcj{
Recent studies have made progress on the attention mechanism of MLLMs.
Researchers develop new attention mechanisms and fine-tune MLLMs to suppress the diminishing attention scores between distant visual cues and output tokens due to Rotary Position Embeddings (RoPE), mitigating hallucinations~\cite{ma2024vista,xing2024concentric}.
Meanwhile, the attention sink phenomenon~\cite{xiao2023attnsink,huang2024opera} has been observed, which refers to tokens receiving significantly higher attention scores than others but providing limited semantic information. 
Distinct from these studies, we explore the attention allocation as well as the feature distribution between different token types rather than the token-level mechanism.
Besides, we establish the correlation between our observation and hallucinations, based on which we introduce a training-free mitigation solution. 
}
\section{Method}
\label{sec:method}

\begin{figure*}[htbp]
    \centering
    \includegraphics[width=0.97\linewidth]{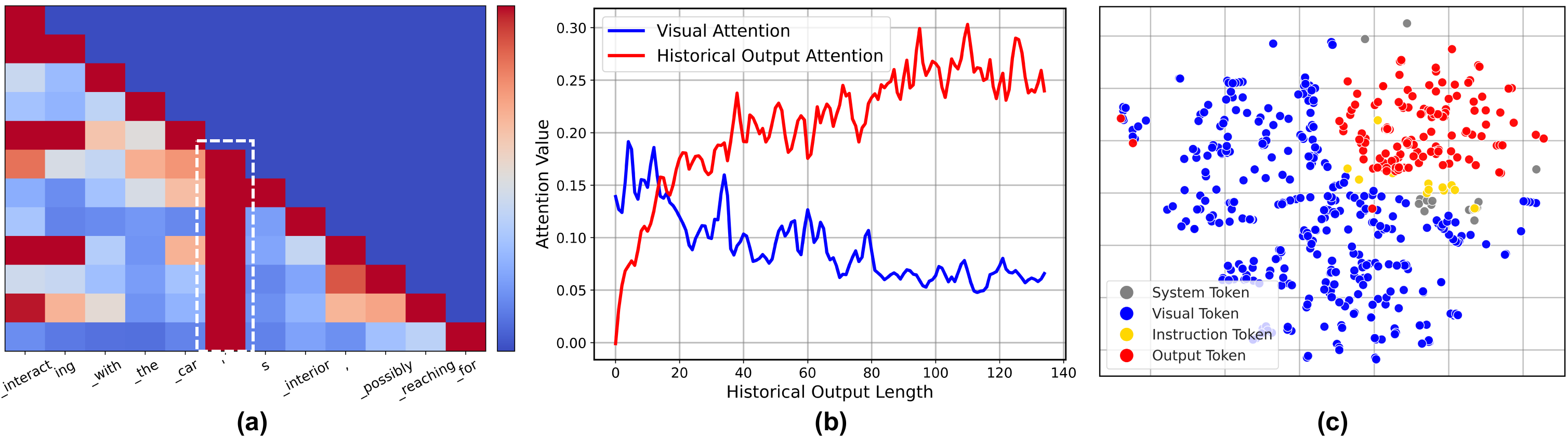}
    \vspace{-3mm}
    \caption{Visualization of our observations obtained with samples from MSCOCO on Qwen2-VL~\cite{wang2024qwen2}. 
                (a) A case of the attention sink phenomenon among output tokens. Red areas represent high attention values. The words corresponding to each token are shown below each column.
                (b) Changes in MLLM's total attention value to historical output tokens and visual tokens.
                (c) Visualization of feature distribution of different tokens in the last decoder layer of the MLLM.
            }
    \label{fig:observations}
    \vspace{-4mm}
\end{figure*}

\subsection{Preliminaries}
In this section, we formulate the MLLM decoding process to facilitate an easier understanding of the proposed attention reallocation method.

\noindent \textbf{Input organization.} 
Most existing MLLMs transform the input contents into token sequences and organize them in the order of system tokens, visual tokens, and instruction tokens.
In the auto-regressive decoding process, models concatenate the historically generated output tokens to the post of the initial input token sequence. Thus, the input sequence of the model can be denoted as $[X_s, X_v, X_i, X_o]$ except for the generation of the first token.

\noindent \textbf{Forward process}. 
The input token sequence is fed through multiple transformer layers to obtain the final hidden states.
For a given input sequence with $n$ tokens $\mathbf{X}=[X_s, X_v, X_i, X_o]$ to a specific layer, the model generates the output hidden states as follows:
\begingroup
\small
\begin{equation}
    h = A f_V(\mathbf{X}), A = \text{Softmax}(\frac{f_Q(\mathbf{X})f_K{(\mathbf{X}})^T}{\sqrt{d_k}}),
\end{equation}
\endgroup
where $f_Q$, $f_K$, and $f_V$ are the projection matrices for the queries, keys, and values, $d_k$ denotes the embedding dimensions and $h$ represents the output hidden states. The attention matrix $A\in \mathbb{R}^{n\times n}$ can be decomposed into matrices for each token type: $A=[A_s, A_v, A_i, A_o]$. Similarly, the value matrix $V=f_V(X)\in \mathbb{R}^{n\times d_k}$ can also be decomposed based on token type: $V=[V_s, V_v, V_i, V_o]^T$. Therefore, the output hidden states can be reformulated as:
\begingroup
\small
\begin{equation}
    h = A_s V_s + A_v V_v + A_i V_i + A_o V_o.
    \label{eq:hidden states per token type}
\end{equation}
\endgroup
Based on~\cref{eq:hidden states per token type}, we can regard the attention to different token types as the weights of tokens during the feature mixing for the output feature, which enables MLLMs to gather information from different token representations.

\textbf{Next token prediction.} The output hidden states of the last layer $h$ are projected using a vocabulary head $\mathcal{H}$ to obtain predicted logits for the next token: 
\begingroup
\small
\begin{equation}
    logits = \text{Softmax}(\mathcal{H}(h)),
    \label{eq:output logits}
\end{equation}
\endgroup
where $logits\in \mathbb{R}^{v}$ and $v$ is the vocabulary set size.

\subsection{Observations}
\label{sec:observations}
\tcj{
One of the major origins of hallucinations in MLLMs lies in their over-reliance on language priors~\cite{leng2024vcd,yue2024lessismore}.
However, how this over-reliance manifests during the decoding process and how it contributes to hallucinated responses remain under-explored. 
In this section, we conduct in-depth exploratory experiments from the perspective of the self-attention layers and token types, considering that MLLM decoding involves various types of tokens. Our observations are as follows.
}



\textbf{MLLM's unreasonable attention allocation gradually amplifies language prior reliance, which leads to output token dominated features.}
\tcj{
As evidenced by~\cref{eq:hidden states per token type}, the output hidden states are obtained by weighing features of different token types using attention values. 
Therefore, we visualize the changes in MLLM's attention to different token types as the length of the historical output token increases.}
\tcj{Specifically, for each generated token, we sum the post-softmax attention values received by each token type and average these sums across all MLLM layers to obtain the results.
As can be concluded from sub-figure (b) of~\cref{fig:observations}, the attention allocated for historical output tokens gradually rises, while visual attention declines, i.e., $A_o$ is increasing and $A_v$ is decreasing in~\cref{eq:hidden states per token type}.
Therefore, this attention allocation pattern reveals that the MLLM gradually relies more on language priors, and the features are gradually dominated by output tokens, especially when the output sequence gets longer.
}


\textbf{The evident distribution gap between different token types translates output token dominance into hallucinated responses.}
\tcj{
To further bridge the correlation between features dominated by historical output tokens and hallucinations, we visualize and analyze the feature distributions of different token types in the last decoder layer of the MLLM.
As can be observed from sub-figure (c) in~\cref{fig:observations}, there exists a significant distribution gap between visual tokens and textual tokens (including system, instruction, and output tokens). 
Since the decoded features are gradually dominated by historical output tokens, the next token tends to distribute closer to textual token embeddings while deviating from visual features. 
When decoding with the visually ungrounded features, the logits of next-token predictions become susceptible to statistical biases from the LLM pretraining process and tend to yield textually coherent but visually irrelevant tokens (e.g., inferring ``people'' when only a car is visible), resulting in hallucinated responses.
}

\begin{figure*}[htbp]
    \centering
    \includegraphics[width=\linewidth]{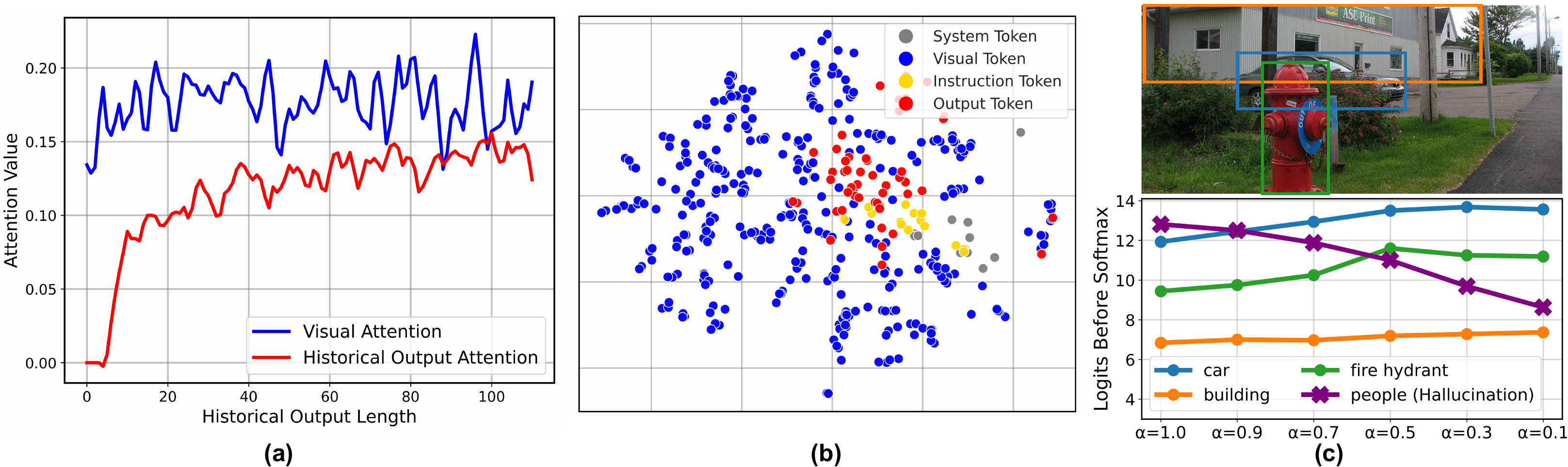}
    \vspace{-7mm}
    \caption{
            (a) Changes in MLLM's total attention value to historical output tokens and visual tokens when applying our AttnReal.
            (b) Feature distribution of different tokens in the last decoder layer of the MLLM when applying our AttnReal.
            (c) The upper part shows the image provided to the MLLM, in which the ground-truth objects are highlighted with different colors. The lower part demonstrates the MLLM's predicted logits for various tokens at different intensities of our AttnReal. 
            }
    \label{fig:method-using-AttnReal}
    \vspace{-3mm}
\end{figure*}

\subsection{Attention Reallocation}
We have revealed that unreasonable attention allocation is an important origin of MLLM hallucinations in~\cref{sec:observations}. 
Therefore, in this section, we propose the attention reallocation (\textbf{AttnReal}) approach with almost zero additional cost to effectively mitigate MLLM hallucinations and improve the consistency of the predicted results with the input visual evidence.
Specifically, AttnReal comprises two steps: attention recycling and attention allocation.

\noindent \textbf{\tcj{Step 1:} attention recycling.}
\tcj{
To avoid MLLMs from relying unreasonably heavily on historical output tokens during the decoding process, we first recycle the attention allocated to output tokens in self-attention layers.
According to the attention sink phenomenon~\cite{xiao2023attnsink,huang2024opera}, there exist some tokens that receive significantly high attention scores but provide limited semantic information among the output tokens (visualized in subfigure (a) of~\cref{fig:observations}).
Based on this phenomenon, we focus on recycling attention from these attention sinks to suppress the dominance of output tokens.
}
Denote the current input token sequence as $[X_s, X_v, X_i, X_o]$ and the corresponding attention map after softmax as $[A_s, A_v, A_i, A_o]$.
We identify the attention sinks in the output tokens $X_o$ by setting a sink threshold $\mathcal{T}$. The identified attention sink set is as follows:
\begingroup
\small
\begin{equation} 
    S_{sink} =  \{x_{j} \mid a_j > \mathcal{T} \cdot \frac{1}{n}\}, \text{ for } x_{j} \in X_{o}, \, j = 1, \dots, n_o,
    \label{eq:identify_sinks}
\end{equation} 
\endgroup
where $a_j \in A_o$ is the corresponding attention score to $x_j$. $n$ denotes the input sequence length, and $n_o$ is the number of output tokens. This equation defines attention sinks to be output tokens receiving more than $\mathcal{T}$ times the average attention score, i.e., $\frac{1}{n}$. 
Next, we introduce a down-scaling factor $\alpha$ for attention recycling, whose value determines the intensity of AttnReal. Specifically, the attention scores $a_j \in A_o$ are updated as follows:
\begingroup
\begin{equation}
    a'_{j} = 
        \begin{cases}
        \alpha \cdot a_{j}, & \text{if } x_{j} \in S_{\text{sink}}, \\
        a_{j}, & \text{if } x_{j} \notin S_{\text{sink}},
        \end{cases}
    \quad \text{for } j = 1, \dots, n_o.
    \label{eq:recycle_attn}
\end{equation}
\endgroup
\textbf{\tcj{Step 2:} attention allocation.}
To utilize the recycled attention to effectively reduce MLLM's hallucinations, we allocate it to visual tokens, enhancing the alignment between the output hidden states and visual features. Denote the reduction of attention to output tokens as $\Delta_{A_o}$ and the number of visual tokens as $n_v$. 
We uniformly allocate the attention to visual tokens $a_k \in A_v$, which can be formulated as:
\begingroup
\small
\begin{equation}
    a'_k = a_k + \frac{\Delta_{A_o}}{n_v}, \, \quad k = 1, \dots, n_v.
    \label{eq:allocate_attn}
\end{equation}
\endgroup
This allocation process increases the MLLM's awareness of the visual information and yields output hidden states better aligned with the visual embeddings.
Overall, AttnReal reallocates the attention values, contributing to better output hidden states and predicted logits, ultimately reducing hallucinations in MLLMs.


\subsection{Effectiveness of Attention Reallocation}
To examine whether AttnReal works as expected, we visualize the attention value and feature distribution of different token types when AttnReal is applied.
As shown in sub-figure (a) of~\cref{fig:method-using-AttnReal}, MLLM's attention to the visual tokens
\tcj{remains broadly stable with fluctuations,}
while the attention to output tokens grows slightly but stays lower than the visual tokens.
Sub-figure (b) demonstrates that by applying AttnReal, the MLLM achieves better alignment between output features and input visual features, 
\tcj{contributing to predictions that are more grounded in the visual embeddings.}

Furthermore, we compare the changes of predicted logits at varying down-scaling factors $\alpha$ in sub-figure (c) of~\cref{fig:method-using-AttnReal} to provide \tcj{further} evidence about how AttnReal reduces hallucinations.
Specifically, given the upper image of sub-figure (c), the MLLM is asked to generate a description. The results of $\alpha=1.0$ in the lower part of sub-figure (c) correspond to the baseline, in which the hallucination token ``people'' achieves the highest logits and is generated, although there does not exist any person in the image.
As we increase the intensity of AttnReal, i.e., reduce the value of $\alpha$, the logits of ``people'' gradually decrease, and the logits of ground-truth tokens like ``car'' and ``building'' increase.
When $\alpha=0.7$, the logits of ``car'' surpass that of ``people'' and the hallucination is eliminated.

\section{Experiments}
\label{sec:exps}

\subsection{Setup}

\begin{table*}[ht]

\caption{
    CHAIR evaluation results on five open-source MLLMs. CHAIR$_\text{S}$ and CHAIR$_\text{I}$ (lower is better) reflect the degree of hallucination. The F1 score (higher is better) represents the comprehensive consideration of accuracy and recall. 
    In \textit{\textbf{CHAIR-aligned}} lines, we align CHAIR$_\text{S}$ and CHAIR$_\text{I}$ with the best hallucination performance of other methods and focus on the F1 score comparison. In \textit{\textbf{F1-aligned}} lines, we align our F1 score with the highest F1 of other methods and emphasize the improvements in CHAIR metrics. ``-'' means the corresponding method is not implemented in this model.
    }
    \vspace{-2.5mm}
\label{tab: chair-5model}
\resizebox{\linewidth}{!}{%
\setlength{\tabcolsep}{1.9mm}
\begin{tabular}{lcccccccccc}
\toprule
\multirow{2}{*}{Methods} & \multirow{2}{*}{Decoding Strategy} &
\multicolumn{3}{c}{LLaVA-1.5-7B} & \multicolumn{3}{c}{LLaVA-1.5-13B} & \multicolumn{3}{c}{MiniGPT-4} \\ 
\cmidrule(lr){3-5} \cmidrule(lr){6-8} \cmidrule(lr){9-11}
 &  & 
  CHAIR$_\text{S}\downarrow$ &  CHAIR$_\text{I}\downarrow$ &  F1 $\uparrow$ &
  CHAIR$_\text{S}\downarrow$ &  CHAIR$_\text{I}\downarrow$ &  F1 $\uparrow$ &
  CHAIR$_\text{S}\downarrow$ &  CHAIR$_\text{I}\downarrow$ &  F1 $\uparrow$ \\
\cmidrule(lr){1-11}

Greedy           & \multirow{5}{*}{Greedy Decoding}  & 47.0 & 14.3 & 77.1 & 44.8 & 12.8 & 77.9  & 31.0  & 9.6  & 70.6  \\
DoLa~\cite{chuang2023dola}             &                                   & 46.8  & 14.3  & 76.9  & -  & -  & -  & 29.2  & 10.1  & 71.5  \\
PAI~\cite{liu2024pai}              &                                   & 29.6  & 7.3  & 76.0  & 35.6  & 9.2  & 77.9  & 24.4  & 9.0  & 70.9  \\
\rowcolor{blue!10}\textbf{Ours (CHAIR-aligned)}&              & 30.6 & 7.0 & \textbf{78.3 \small{\red{($\uparrow$1.2)}}} & 36.6 & 8.7 & \textbf{79.7 \small{\red{($\uparrow$1.8)}}} & 25.6 & 7.7 & 70.9 \\
\rowcolor{green!10}\textbf{Ours (F1-aligned)}            &     & \textbf{23.0 \small{\red{($\downarrow$6.6)}}} & \textbf{5.3 \small{\red{($\downarrow$1.7)}}} & 77.3 & \textbf{24.6 \small{\red{($\downarrow$11.0)}}} & \textbf{5.0 \small{\red{($\downarrow$4.2)}}} & 78.0 & 27.8 & \textbf{8.2 \small{\red{($\downarrow$0.8)}}} & 71.5 \\
\cmidrule(lr){1-11}
Beam Search      & \multirow{6}{*}{Beam Search}      & 51.8  & 15.5  & 77.0  & 49.0  & 13.4  & 78.3  & 34.4  & 10.9  & 70.4  \\
OPERA~\cite{huang2024opera}            &                                   & 48.6  & 14.1  & 76.9  & 40.4  & 12.4  & 76.8  & 28.6  & 10.1  & 69.6  \\
VCD-Beam~\cite{leng2024vcd}         &                                   & 52.6  & 16.0  & 74.8  & 48.4  & 14.5  & 75.0  & 32.4  & 9.8  & 69.8  \\
PAI-Beam~\cite{liu2024pai}         &                                   & 35.0  & 10.2  & 75.6  & 40.4  & 12.6  & 77.0  & 24.0  & 8.7  & 69.8  \\
\rowcolor{blue!10}\textbf{Ours-Beam (CHAIR-aligned)}    &    & 36.6 & 9.8 & \textbf{78.4 \small{\red{($\uparrow$1.4)}}} & 40.0 & 9.6 & \textbf{79.5 \small{\red{($\uparrow$1.2)}}} & 24.2 & 6.7 & \textbf{71.4 \small{\red{($\uparrow$1.0)}}} \\
\rowcolor{green!10}\textbf{Ours-Beam (F1-aligned)}       &    & \textbf{21.6 \small{\red{($\downarrow$13.4)}}} & \textbf{6.4 \small{\red{($\downarrow$3.8)}}} & 77.4 & \textbf{27.8 \small{\red{($\downarrow$12.6)}}} & \textbf{6.5 \small{\red{($\downarrow$6.1)}}} & 78.5 & \textbf{17.2 \small{\red{($\downarrow$6.8)}}} & \textbf{5.5 \small{\red{($\downarrow$3.2)}}} & 70.3 \\
\cmidrule(lr){1-11}
Nucleus Sampling & \multirow{5}{*}{Nucleus Sampling} & 52.8  & 17.3  & 72.7  & 54.2  & 15.6  & 73.8  & 34.6  & 11.0  & 69.5  \\
VCD~\cite{leng2024vcd}              &                                   & 48.4  & 15.1  & 74.3  & 50.0  & 14.5  & 76.0  & 32.0  & 10.3  & 68.8  \\
PAI-Nucleus~\cite{liu2024pai}      &                                   & 39.2  & 10.8  & 74.0  & 41.0  & 12.8  & 76.1  & 23.2  & 9.2  & 70.9  \\
\rowcolor{blue!10}\textbf{Ours-Nucleus (CHAIR-aligned)}    &     & 42.0 & 10.6 & \textbf{75.0 \small{\red{($\uparrow$0.7)}}} & 42.6 & 10.1 & \textbf{77.6 \small{\red{($\uparrow$1.5)}}} & 25.4 & 7.6 & 70.0 \\
\rowcolor{green!10}\textbf{Ours-Nucleus (F1-aligned)}    &        & 42.2 & \textbf{9.3 \small{\red{($\downarrow$1.5)}}}  & 74.8 & \textbf{37.2 \small{\red{($\downarrow$3.8)}}} & \textbf{7.9 \small{\red{($\downarrow$4.9)}}} & 76.2 & 26.0 & 9.9 & 70.5 \\

\bottomrule
\end{tabular}
}

\vspace{1mm}

\resizebox{\linewidth}{!}{%
\setlength{\tabcolsep}{4.5mm}
\begin{tabular}{lccccccc}
\toprule
\multirow{2}{*}{Methods} & \multirow{2}{*}{Decoding Strategy} &
\multicolumn{3}{c}{Shikra} & \multicolumn{3}{c}{mPLUG-Owl2} \\ 
\cmidrule(lr){3-5} \cmidrule(lr){6-8}
 &  & 
  CHAIR$_\text{S}\downarrow$ &  CHAIR$_\text{I}\downarrow$ &  F1 $\uparrow$ &
  CHAIR$_\text{S}\downarrow$ &  CHAIR$_\text{I}\downarrow$ &  F1 $\uparrow$ \\
\cmidrule(lr){1-8}

Greedy           & \multirow{5}{*}{Greedy Decoding}  & 56.6  & 16.4  & 74.5  & 59.0  & 19.3  & 71.1  \\
DoLa~\cite{chuang2023dola}             &                                   & -  & -  & -  & 58.2  & 18.6  & 71.5  \\
PAI~\cite{liu2024pai}              &                                   & 31  & 8.6  & 74.9  & -  & -  & -    \\
\rowcolor{blue!10}\textbf{Ours (CHAIR-aligned)}              &         & 31.2 & 7.6 & \textbf{75.6 \small{\red{($\uparrow$0.7)}}} & 37.0 & 11.1 & \textbf{75.3 \small{\red{($\uparrow$3.8)}}} \\
\rowcolor{green!10}\textbf{Ours (F1-aligned)}             &             & \textbf{24.2 \small{\red{($\downarrow$6.8)}}} & \textbf{5.9 \small{\red{($\downarrow$2.7)}}} & 75.3 & \textbf{19.8 \small{\red{($\downarrow$38.4)}}} & \textbf{6.2 \small{\red{($\downarrow$12.4)}}} & 72.9 \\
\cmidrule(lr){1-8}
Beam Search      & \multirow{6}{*}{Beam Search}      & 57.0  & 15.5  & 75.1  & 58.2  & 18.1  & 73.1    \\
OPERA~\cite{huang2024opera}            &                                   & 38.4  & 12.9  & 73.8  & 50.8  & 16.9  & 72.2    \\
VCD-Beam~\cite{leng2024vcd}         &                                   & -  & -  & -  & 59.6  & 19.9  & 68.8   \\
PAI-Beam~\cite{liu2024pai}         &                                   & 33.8  & 10.0  & 74.5  & -  & -  & -   \\
\rowcolor{blue!10}\textbf{Ours-Beam (CHAIR-aligned)}        &          & 34.2 & 8.1 & \textbf{77.0 \small{\red{($\uparrow$1.9)}}} & 43.0 & 13.6 & \textbf{75.2 \small{\red{($\uparrow$2.1)}}}   \\
\rowcolor{green!10}\textbf{Ours-Beam (F1-aligned)}        &             & \textbf{21.8 \small{\red{($\downarrow$12.0)}}} & \textbf{5.2 \small{\red{($\downarrow$4.8)}}} & 75.1 & \textbf{22.8 \small{\red{($\downarrow$28.0)}}} & \textbf{7.9 \small{\red{($\downarrow$9.0)}}} & 73.2   \\
\cmidrule(lr){1-8}
Nucleus Sampling & \multirow{5}{*}{Nucleus Sampling} & 60.8  & 16.1  & 72.5  & 61.2  & 19.8  & 68.9  \\
VCD~\cite{leng2024vcd}               &                                   & -  & -  & -  & 58.4  & 20.3  & 68.9   \\
PAI-Nucleus~\cite{liu2024pai}       &                                   & 34.2  & 8.9  & 74.1  & -  & -  & -  \\
\rowcolor{blue!10}\textbf{Ours-Nucleus (CHAIR-aligned)}      &         & 36.6 & 7.9 & \textbf{76.6 \small{\red{($\uparrow$2.5)}}} & 36.8 & 11.4 & \textbf{72.3 \small{\red{($\uparrow$3.4)}}}   \\
\rowcolor{green!10}\textbf{Ours-Nucleus (F1-aligned)}     &             & \textbf{33.0 \small{\red{($\downarrow$1.2)}}} & \textbf{7.1 \small{\red{($\downarrow$1.8)}}} & 74.2 & \textbf{28.6 \small{\red{($\downarrow$29.8)}}} & \textbf{9.0 \small{\red{($\downarrow$11.3)}}} & 70.4  \\
\bottomrule
\end{tabular}%
}
\vspace{-4mm}
\end{table*}
\begin{table}[htbp]
\caption{
    CHAIR evaluation results on Qwen2-VL~\cite{wang2024qwen2}. CHAIR$_\text{S}$ and CHAIR$_\text{I}$ reflect the degree of hallucination, and the F1 score represents the comprehensive consideration of accuracy and recall. 
    }
\label{tab: chair-part3}
\vspace{-2mm}
\resizebox{\columnwidth}{!}{%
\begin{tabular}{lcccc}
\toprule
\multirow{2}{*}{Methods} & \multirow{2}{*}{Decoding Strategy}   & \multicolumn{3}{c}{Qwen2-VL}   \\ 
\cmidrule(lr){3-5}
                         &                                  & CHAIR$_\text{S}\downarrow$ & CHAIR$_\text{I}\downarrow$ & F1 $\uparrow$ \\ 
\cmidrule(lr){1-5}

Greedy                   & \multirow{3}{*}{Greedy Decoding} & 30.0         & 8.9         &  77.1        \\
\textbf{Ours {\small (CHAIR-aligned)}}                     &                                  & 25.8        & 7.8        & \textbf{78.3 \small{\red{($\uparrow$1.2)}}}        \\
\textbf{Ours (F1-aligned)}                     &                                  & \textbf{21.0\small{\red{($\downarrow$9.0)}}}        & \textbf{5.9\small{\red{($\downarrow$3.0)}}}        & 77.1        \\
\bottomrule
\end{tabular}%
}
\vspace{-5mm}
\end{table}

\noindent \textbf{Baselines.}
In this paper, we choose six representative MLLM backbones to validate the plug-and-play effectiveness of our proposed AttnReal approach, including LLaVA-1.5-7B~\cite{liu2024visualllava}, LLaVA-1.5-13B, MiniGPT-4~\cite{zhu2023minigpt}, Shikra~\cite{chen2023shikra}, mPLUG-Owl2~\cite{ye2024mplug}, and Qwen2-VL~\cite{wang2024qwen2}. These models are 7B models except for LLaVA-1.5-13B.
Moreover, to show the wide applicability, we apply AttnReal to three decoding strategies: Greedy Decoding, Beam Search, and Nucleus Sampling.
For performance comparisons, we focus on training-free methods for MLLM hallucination mitigation, including DoLa~\cite{chuang2023dola}, OPERA~\cite{huang2024opera}, VCD~\cite{leng2024vcd}, and PAI~\cite{liu2024pai}. 
DoLa is applied to Greedy Decoding by default.
OPERA relies on Beam Search. VCD is based on Nucleus Sampling and can be applied with Beam Search. PAI can be adopted with all three decoding strategies.
All baselines are implemented with the default hyperparameters from their open-source codes.

\begin{figure*}[ht]
    \centering
    \includegraphics[width=0.95\linewidth]{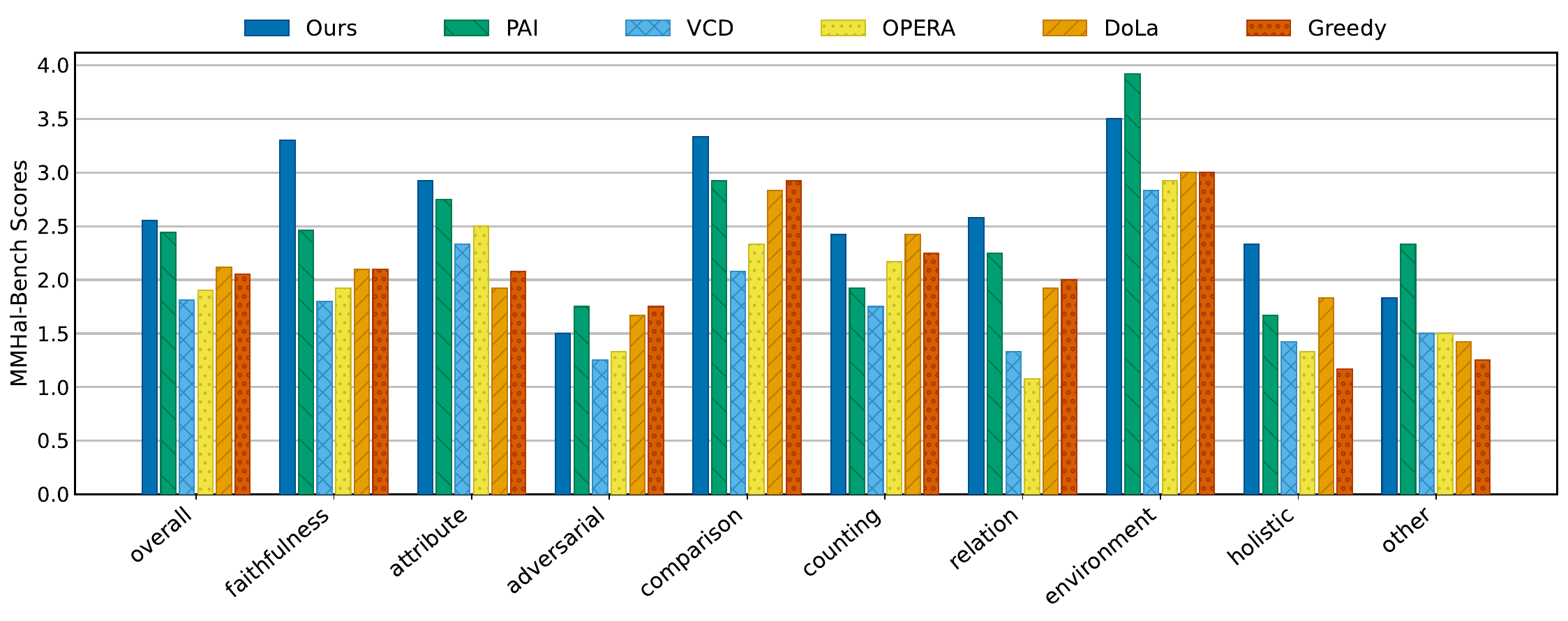}
    \vspace{-4.5mm}
    \caption{Comparative results on the MMHal-Bench and LLaVA-1.5-7B. To all metrics, higher scores mean better performance.}
    \label{fig:mmhal}
    \vspace{-3mm}
\end{figure*}

\noindent \textbf{Implementation details.}
All compared methods are evaluated with the same random seed and decoding strategy settings. For Beam Search, the number of beam candidates is set to 5, and for Nucleus Sampling the common parameters are consistent.
For AttnReal, we set the sink threshold $\mathcal{T}=1$ for all MLLMs by default unless otherwise stated, given that the attention value of most tokens is much smaller than the average due to attention sinks.
Moreover, AttnReal is robust to different $\mathcal{T}$ values, as demonstrated in~\cref{sec:ablation}.

\subsection{Quantitative Results}
In this section, we evaluate the performance of AttnReal with MLLM hallucination evaluation benchmarks and GPT-assisted open evaluation, covering evaluations for various hallucination types and overall performance.

\subsubsection{CHAIR Evaluation}  
\textbf{C}aption \textbf{H}allucination \textbf{A}ssessment with \textbf{I}mage \textbf{R}elevance (CHAIR)~\cite{rohrbach2018chair} is a specifically crafted benchmark to assess object hallucination in the image captioning task. Specifically, CHAIR provides the ground-truth object labels for all samples and evaluates hallucinations in given image descriptions by calculating the ratio of all objects mentioned in the description that are not present in the ground-truth label set.
CHAIR comprises two quantitative metrics, $\text{CHAIR}_\text{S}$ and $\text{CHAIR}_\text{I}$, to assess the sentence-level and instance-level hallucination, respectively. 
Besides hallucination metrics, we also adopt the F1 score, which is decided by both precision and recall, to evaluate the overall performance.
We follow the settings of~\cite{huang2024opera, liu2024pai} to perform CHAIR evaluation on the validation set of MSCOCO 2014~\cite{lin2014mscoco}. Specifically, we randomly sample 500 samples from the validation set and prompt the MLLMs with \texttt{"Please describe this image in detail."} for caption generation, with the \textit{max\_new\_tokens} parameter set to 512.

\noindent \textbf{Results of CHAIR.}
\cref{tab: chair-5model,tab: chair-part3} exhibit the comprehensive results of different methods on 6 MLLMs and 3 decoding strategies. Since our proposed AttnReal achieves a wide-range trade-off between CHAIR metrics and F1 scores, as illustrated in~\cref{fig: figure1}, we show our performances under two settings: 
1) We align the hallucination performance (CHAIR$_\text{S}$ and CHAIR$_\text{I}$) with the lowest CHAIR metrics of existing methods and compare the F1 scores.
2) We also align the F1 performance with the best of other methods and compare the CHAIR metrics.
From~\cref{tab: chair-5model} we can conclude that for most MLLMs and decoding strategies, our proposed method achieves higher F1 scores with aligned CHAIR metrics and yields lower hallucinations with aligned F1, demonstrating its effectiveness and wide applicability.
For Qwen2-VL~\cite{wang2024qwen2}, which has not been integrated by existing studies, we enhance the F1 score from 77.1 to 78.3 with lower hallucination, and reduce the CHAIR$_\text{S}$/CHAIR$_\text{I}$ metrics from 30.0 / 8.9 to 21.0 / 5.9 with aligned F1 score.
More results on the CHAIR benchmark are provided in~\cref{app:chair} of the appendix.

\begin{table*}[ht]
\centering
\caption{Quantitative comparison on GPT-assisted evaluation. The best results are \textbf{bolded}, and the suboptimal results are \underline{underlined}. $\mathcal{C}$ and $\mathcal{D}$ represent for correctness and detailedness, respectively. ``-'' means the corresponding method is not implemented in this model.}
\label{tab: open evaluatoin}
\vspace{-2mm}
\resizebox{\textwidth}{!}{%
\setlength{\tabcolsep}{3.5mm}
\begin{tabular}{lcccccccccccc}
\toprule
\multirow{2}{*}{Method} &
  \multicolumn{2}{c}{LLaVA-1.5-7B} &
  \multicolumn{2}{c}{LLaVA-1.5-13B} &
  \multicolumn{2}{c}{MiniGPT-4} &
  \multicolumn{2}{c}{Shikra} &
  \multicolumn{2}{c}{mPLUG-Owl2} & 
  \multicolumn{2}{c}{Qwen2-VL} \\ 
\cmidrule(lr){2-3} \cmidrule(lr){4-5} \cmidrule(lr){6-7} \cmidrule(lr){8-9} \cmidrule(lr){10-11} \cmidrule(lr){12-13}
 &
  $\mathcal{C}$ &  $\mathcal{D}$ &
  $\mathcal{C}$ &  $\mathcal{D}$ &
  $\mathcal{C}$ &  $\mathcal{D}$ &
  $\mathcal{C}$ &  $\mathcal{D}$ &
  $\mathcal{C}$ &  $\mathcal{D}$ &
  $\mathcal{C}$ &  $\mathcal{D}$ \\ 
\cmidrule(lr){1-13}
Greedy & 3.22 & 3.35 & 3.38 & 3.43 & 2.80  & 3.25 & 2.99 & 3.23 & 2.66 & \underline{3.10} & \underline{4.08} & \underline{3.93}  \\
DoLa~\cite{chuang2023dola}            & 3.25 & \underline{3.41} & - & - & 2.81 & 3.21 & -    & -    & \underline{2.67} & 3.08 & - & -  \\
OPERA~\cite{huang2024opera}           & \underline{3.31} & \underline{3.41} & 3.36 & 3.37 & 3.03 & 3.22 & 3.02 & 3.14 & 2.65 & 3.07 & - & - \\
VCD~\cite{leng2024vcd}             & 3.15 & 3.32 & 3.18 & 3.34 & 2.88 & 3.22 & -    & -    & 2.56 & 3.00 & - & -    \\
PAI~\cite{liu2024pai}             & 3.24 & 3.29 & \underline{3.40}  & \underline{3.53} & \textbf{3.13} & \textbf{3.36} & \underline{3.13} & \underline{3.28} & -    & -   & - & -   \\
\textbf{Ours}            & \textbf{3.40}  & \textbf{3.44} & \textbf{3.47} & \textbf{3.55} & \underline{3.07} & \underline{3.35} & \textbf{3.14} & \textbf{3.36} & \textbf{3.04} & \textbf{3.36} & \textbf{4.17} & \textbf{3.97} \\
\bottomrule
\end{tabular}%
}
\vspace{-3mm}
\end{table*}

\subsubsection{MMHal-Bench}
MMHal-Bench~\cite{sun2023aligning_train_mmhal} is a challenging benchmark to evaluate the model's comprehensive performance across various hallucination types. It consists of 96 samples that are evenly distributed across 8 different hallucination types, including object attributes, adversarial objects, comparisons, counting, spatial relations, environment, holistic descriptions, and other hallucinations. 
For the model's responses, MMHal-Bench uses GPT-4 to compare them with carefully crafted ground-truth answers and rates the output for each sample between 0-6 depending on three aspects: informativeness, hallucination, and logical analysis.
The evaluation results include the overall score and hallucination rate on all samples, as well as ratings for each hallucination type. We convert the hallucination rate into a faithfulness rate and scale it to a score between 0-6 as an overall faithfulness score to display together with other ratings. 

\noindent \textbf{Results of MMHal-Bench.}
In~\cref{fig:mmhal}, we show the comparative results of various methods on the MMHal-Bench and LLaVA-1.5-7B. Our proposed AttnReal method achieves the best overall performance and significantly enhances the faithfulness scores compared to existing methods.
Besides, AttnReal receives the highest ratings on 5 out of 8 hallucination types and gains improvement on 7 of 8 hallucination types compared to Greedy Decoding, exhibiting strong competitiveness.
Please refer to~\cref{app:mmhal} in the appendix for the results of other MLLMs on MMHal-Bench.

\subsubsection{GPT-assisted Evaluation}
We further rely on GPT-4o, a strong multi-modal assistant, to conduct the open evaluation of MLLMs' output quality.
Compared to CHAIR~\cite{rohrbach2018chair}, GPT-assisted evaluation provides more assessments beyond object hallucination. 
We follow~\cite{huang2024opera,liu2024pai,leng2024vcd} to randomly select 500 samples from the validation set of MSCOCO and ask MLLMs to generate corresponding descriptions. Then we provide the responses along with the images to GPT-4o, prompting it to rate the correctness ($\mathcal{C}$) and detailedness ($\mathcal{D}$) of each output from 0-5 respectively. Please refer to~\cref{app:gpt prompt} in the appendix for detailed prompts used in this section.

\noindent \textbf{Results of GPT-assisted evaluation.}
Quantitative comparison of various methods on GPT-assisted evaluation is provided in~\cref{tab: open evaluatoin}. 
Our proposed method yields both the best correctness scores and the best detailedness scores on most MLLMs. This means that when integrated with our proposed method, MLLMs can generate more precise responses without sacrificing the details in the input images.

\subsection{Ablation Study}
\label{sec:ablation}
Our proposed AttnReal first identifies and suppresses attention sinks in the historical output tokens and then reallocates the reduced attention to visual tokens. 
This process involves two hyper-parameters in total: the sink threshold $\mathcal{T}$ and down-scaling factor $\alpha$.
We select LLaVA-1.5-7B as the representative MLLM to examine the impact of different hyper-parameters on the performance. The ablation study is conducted on the CHAIR benchmark, and the generated responses are evaluated with both CHAIR metrics and the F1 score.
For a given sink threshold $\mathcal{T}$, we can adjust the intensity of AttnReal, achieving a wide-range trade-off between response faithfulness and overall performance through $\alpha$, as illustrated in~\cref{fig: figure1}.

\begin{figure}[t]
    \centering
    \includegraphics[width=\linewidth]{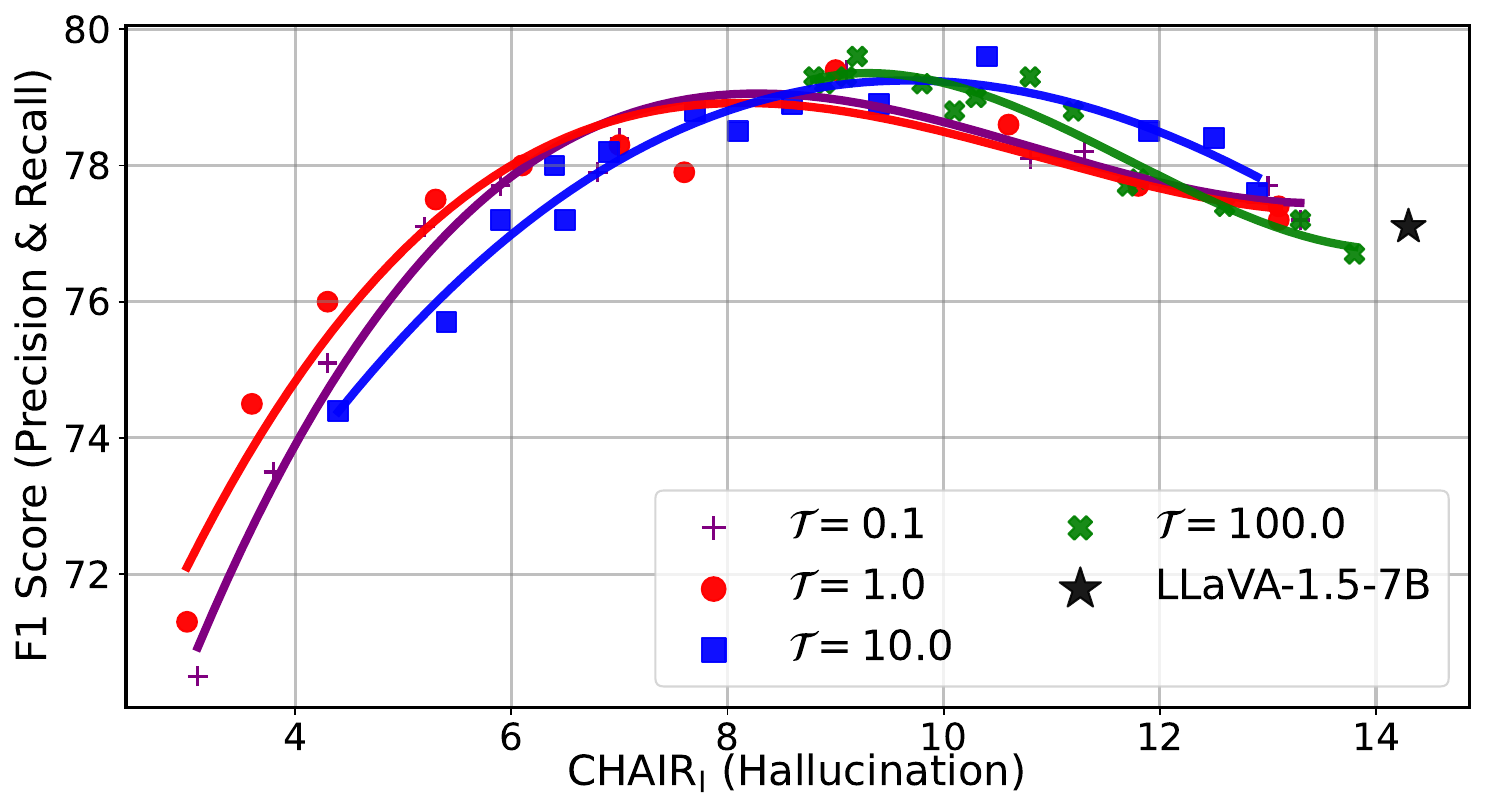}
    \vspace{-8mm}
    \caption{Ablation study on LLaVA-1.5-7B about the effect of different sink thresholds $\mathcal{T}$ on the performance.}
    \label{fig:ablation-threshold}
    \vspace{-4mm}
\end{figure}

We further explore the impact of different $\mathcal{T}$ values on the performance.
Specifically, we selected four values with large differences to demonstrate the robustness of AttnReal to sink threshold configurations, i.e., $\mathcal{T}=[0.1,1,10,100]$.
As can be concluded from~\cref{fig:ablation-threshold}, the results of different settings show a consistent tendency. As $\alpha$ decreases from 1, the CHAIR metrics gradually reduce from the baseline, representing decreased hallucinations.
The F1 score first rises with the intensity of AttnReal and begins to drop when the hallucinations have been significantly suppressed. This change is because the F1 score is determined by both precision and recall. With lower intensities of AttnReal, there is an evident improvement in precision, which leads to a higher F1 score. 
However, when the intensity of AttnReal is high, there are fewer hallucinations, and it is difficult to further improve the precision. At this point, the MLLM tends to focus on the most prominent visual features and may discard some tokens with less certainty, which leads to a lower recall and a subsequent decline in the F1 score.

\section{Conclusion}
\label{sec:conclusion}

In this paper, we observe that unreasonable attention allocation in MLLMs causes features to be dominated by historical output tokens, leading to hallucinated responses due to feature gaps between visual and output tokens. 
To address this, we propose a nearly zero-cost method called attention reallocation (AttnReal), reducing excessive attention to output tokens and encouraging MLLM to rely more on visual evidence.
By adjusting the down-scaling factor of AttnReal, we can balance response faithfulness and overall performance as needed.
Evaluations across various MLLMs and decoding strategies verify the effectiveness of AttnReal.
The limitation of our work is that AttnReal relies on historical output tokens and is not well-suited to addressing hallucinations in tasks with particularly short responses.
Besides, it is worth investigating how to extend the zero-cost hallucination mitigation solution to more tasks and models.

\section*{Impact Statement}
This paper introduces AttnReal, a training-free approach to effectively mitigate hallucinations in Multimodal Large Language Models (MLLMs) with nearly zero extra overhead.

By redistributing attention values during the decoding process, AttnReal can significantly enhance MLLMs' response faithfulness and reliability in safety-critical applications such as autonomous driving and healthcare, where hallucinated responses could lead to severe consequences.

Besides, AttnReal requires no additional training data, avoids costly fine-tuning, and eliminates extra inference latency introduced by methods of multiple decoding steps. This efficiency reduces computational resource consumption and lowers the carbon footprint associated with training and deploying MLLMs.

\bibliography{main}
\bibliographystyle{icml2025}

\newpage
\appendix
\onecolumn

\section{More Experimental Results}
\label{sec:more results}

\subsection{CHAIR Benchmark}
\label{app:chair}
\begin{figure*}[htbp]
    \centering
    \includegraphics[width=\linewidth]{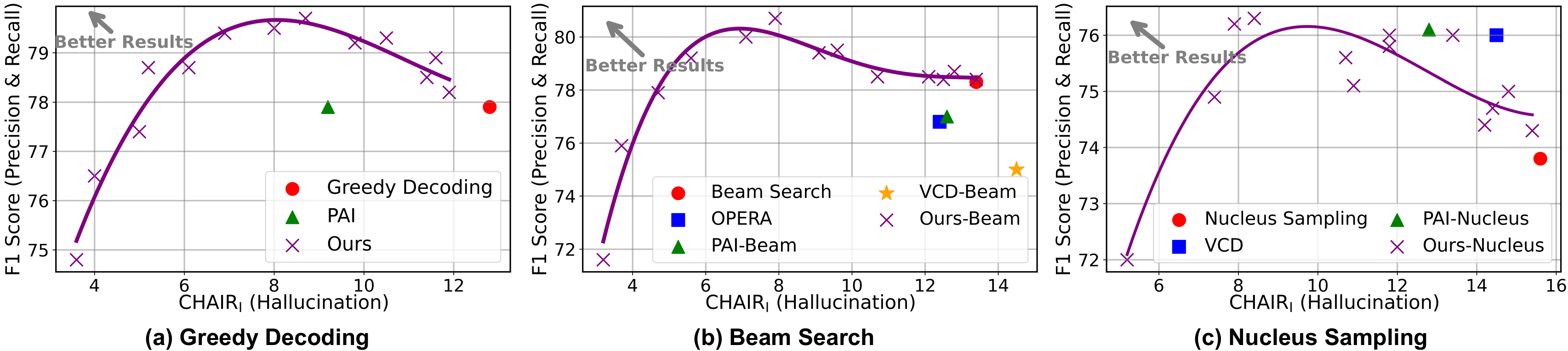}
    \vspace{-6mm}
    \caption{
    Performance comparison for various training-free methods to mitigate MLLM hallucinations on LLaVA-1.5-13B and the CHAIR benchmark using three decoding strategies.
    Lower CHAIR$_\text{I}$ represents fewer hallucinations. Higher F1 scores mean better overall performance of the responses. The curve in each sub-figure is obtained by adjusting the intensity of AttnReal.}
    \label{fig:chair-llava13b}
    \vspace{-2mm}
\end{figure*}

\begin{figure*}[htbp]
    \centering
    \includegraphics[width=\linewidth]{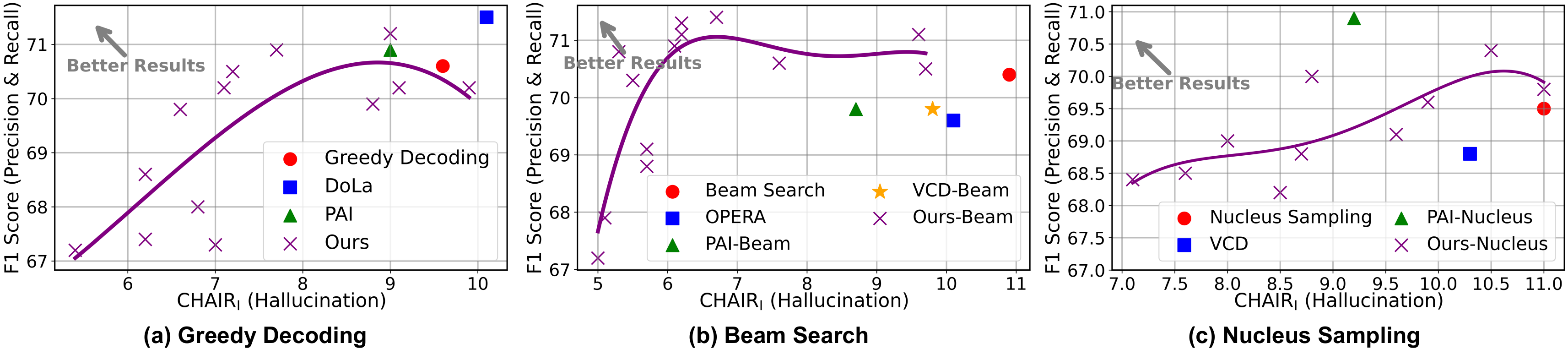}
    \vspace{-6mm}
    \caption{
    Performance comparison for various training-free methods to mitigate MLLM hallucinations on MiniGPT-4 and the CHAIR benchmark using three decoding strategies.
    Lower CHAIR$_\text{I}$ represents fewer hallucinations. Higher F1 scores mean better overall performance of the responses. The curve in each sub-figure is obtained by adjusting the intensity of AttnReal.}
    \label{fig:chair-minigpt4}
    \vspace{-2mm}
\end{figure*}

\begin{figure*}[htbp]
    \centering
    \includegraphics[width=\linewidth]{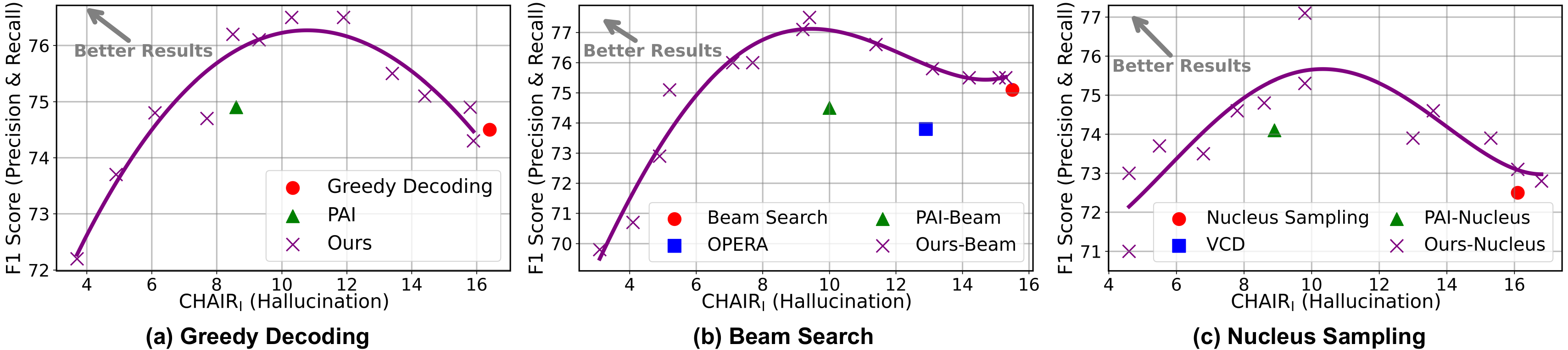}
    \vspace{-6mm}
    \caption{
    Performance comparison for various training-free methods to mitigate MLLM hallucinations on Shikra and the CHAIR benchmark using three decoding strategies.
    Lower CHAIR$_\text{I}$ represents fewer hallucinations. Higher F1 scores mean better overall performance of the responses. The curve in each sub-figure is obtained by adjusting the intensity of AttnReal.}
    \label{fig:chair-shikra}
    \vspace{-2mm}
\end{figure*}

\begin{figure*}[htbp]
    \centering
    \includegraphics[width=\linewidth]{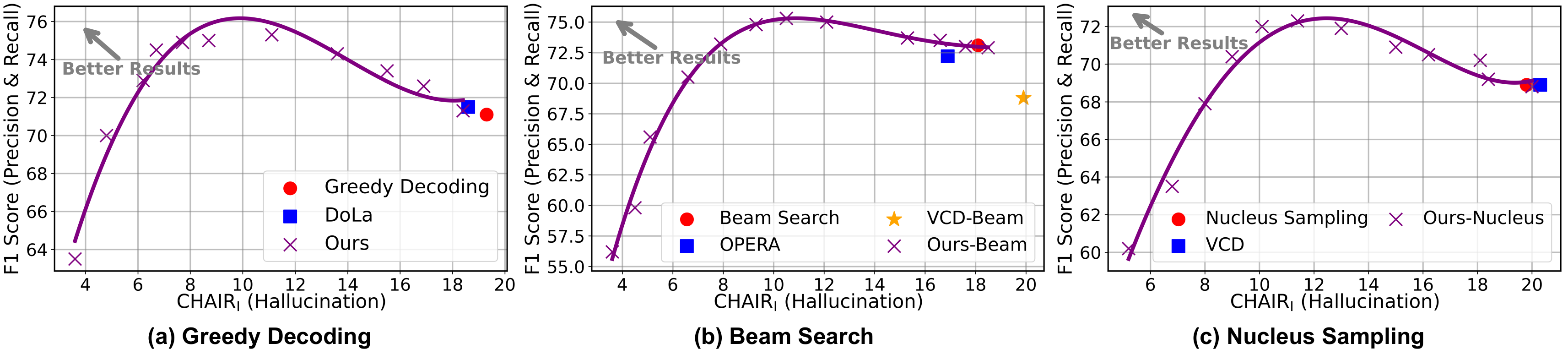}
    \vspace{-6mm}
    \caption{
    Performance comparison for various training-free methods to mitigate MLLM hallucinations on mPLUG-Owl2 and the CHAIR benchmark using three decoding strategies.
    Lower CHAIR$_\text{I}$ represents fewer hallucinations. Higher F1 scores mean better overall performance of the responses. The curve in each sub-figure is obtained by adjusting the intensity of AttnReal.}
    \label{fig:chair-mplug}
    \vspace{-2mm}
\end{figure*}

\begin{figure}[htbp]
\vspace{-5mm}
\begin{multicols}{2}
    \begin{subfigure}
        \centering
        \includegraphics[width=0.9\linewidth]{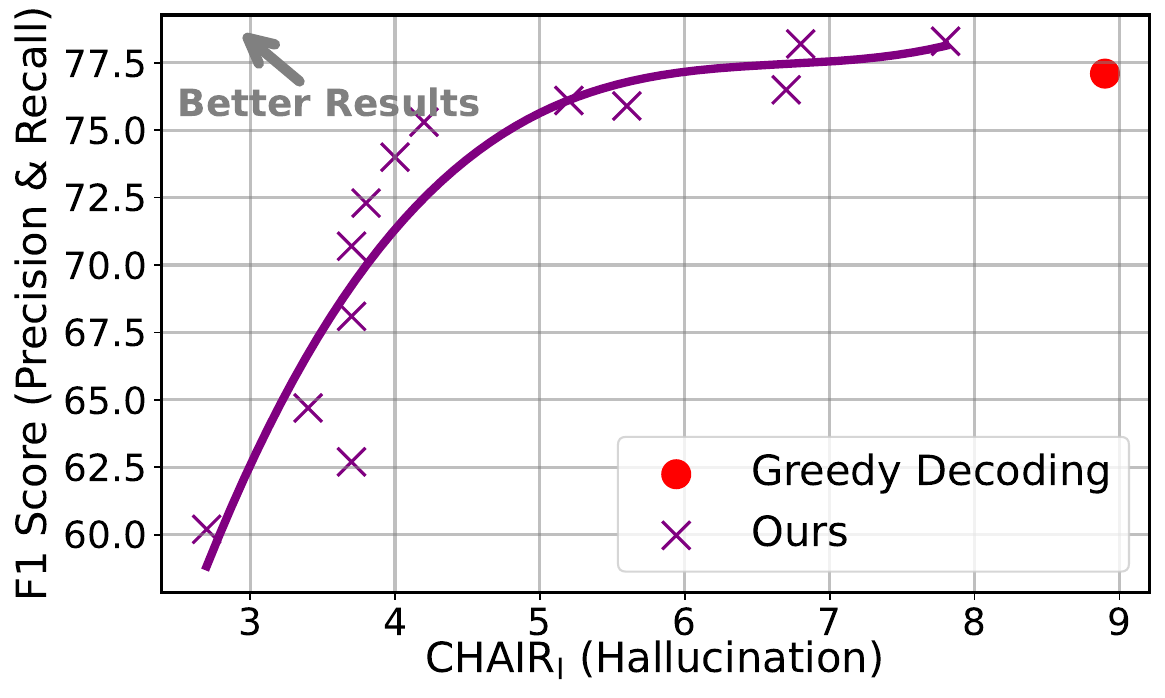}
        \vspace{-3mm}
        \caption{Performance comparison for AttnReal and the baseline Greedy Decoding on Qwen2-VL and the CHAIR benchmark. 
        }
        \label{fig:chair-qwen2}
    \end{subfigure}\break
    
    \begin{subfigure}
        \centering
        \includegraphics[width=\linewidth]{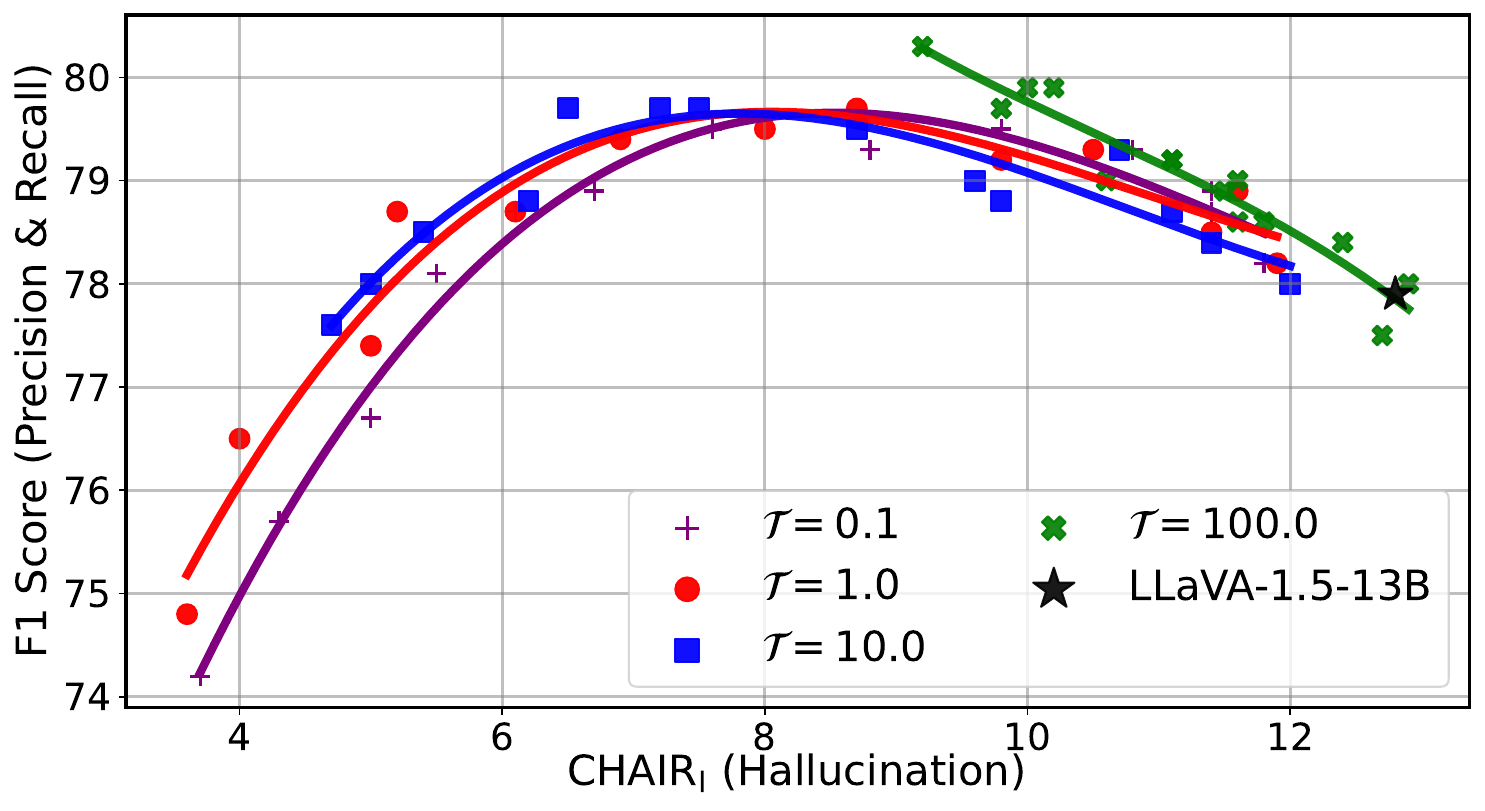}
        \vspace{-6mm}
        \caption{Ablation study on LLaVA-1.5-13B about the effect of different sink thresholds $\mathcal{T}$ on the performance.}
        \label{fig:ablation-threshold-13b}
    \end{subfigure}
\end{multicols}
\vspace{-6mm}
\end{figure}

In this section, we provide more comparative performances on the CHAIR benchmark. 
Specifically, we apply our proposed AttnReal approach to various MLLMs  (including LaVA-1.5-13B, MiniGPT-4, Shikra, mPLUG-Owl2, and Qwen2-VL) and decoding strategies using different down-scaling factors $\alpha$ and compare the performance with existing training-free methods designed for mitigating MLLM hallucinations. The results are illustrated in~\cref{fig:chair-llava13b,fig:chair-minigpt4,fig:chair-shikra,fig:chair-mplug,fig:chair-qwen2}.
After applying AttnReal, hallucinations of the model are significantly reduced and an increase in the F1 score can be obtained under a wide range of values of $\alpha$ in most settings.
Compared to existing methods, AttnReal achieves better performance on most models and decoding approaches, yielding results that outperform the state-of-the-art methods on both CHAIR metrics and the F1 score.
These experimental results further validate the wide applicability of AttnReal on MLLMs of different scales and structures.

\subsection{Ablation Study about Attention Sink Threshold $\mathcal{T}$ on Larger Model}
To explore the robustness of AttnReal to the attention sink threshold $\mathcal{T}$, we choose LLaVA-1.5-13B, which is of larger scale than LLaVA-1.5-7B explored in~\cref{sec:ablation}, for further validation.
As can be concluded from~\cref{fig:ablation-threshold-13b}, although the values of T vary considerably, i.e., $\mathcal{T}=[0.1, 1, 10, 100]$, the results of different settings fall on similar trajectories as the intensity of AttnReal enhances and hallucinations decrease.
Specifically, at lower AttnReal intensities, hallucinations are reduced from baseline, and F1 scores are improved because of the significant increase in precision. When the AttnReal intensity becomes higher, the hallucinations continue to decrease, but the F1 score decreases because the precision cannot be further improved, and the model discards uncertain contents, resulting in lower recall.
This supplementary ablation study demonstrates that AttnReal is also robust to attention sink threshold settings on larger models.

\subsection{Ablation Study about Random Seeds}
\label{app:random seed}
\begin{figure}[htbp]
    \centering
    \includegraphics[width=\linewidth]{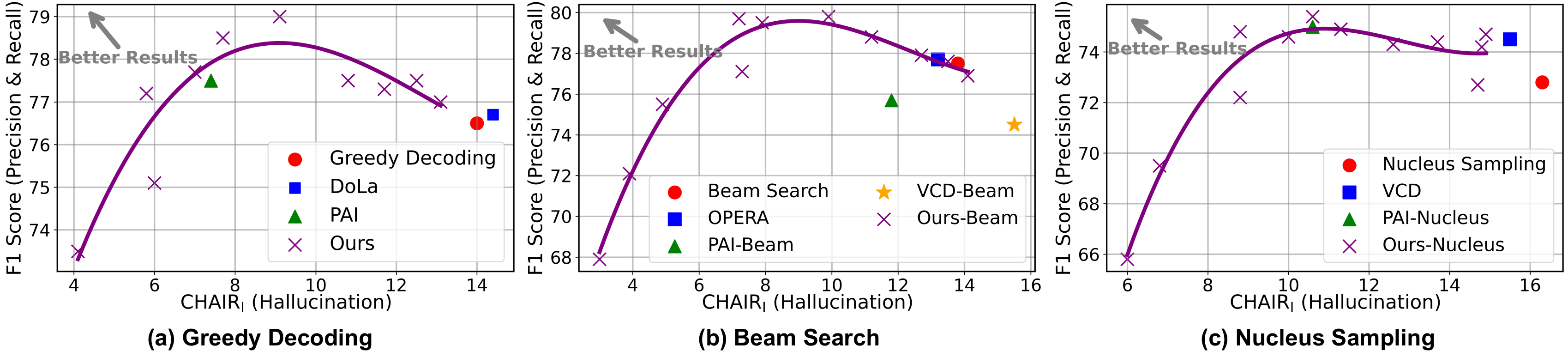}
    \vspace{-2mm}
    \caption{Ablation study on LLaVA-1.5-7B using different random seeds  ($seed$ = 48).}
    \label{fig:ablation-llava-7b-seed48}
\end{figure}

\begin{figure}[htbp]
    \centering
    \includegraphics[width=\linewidth]{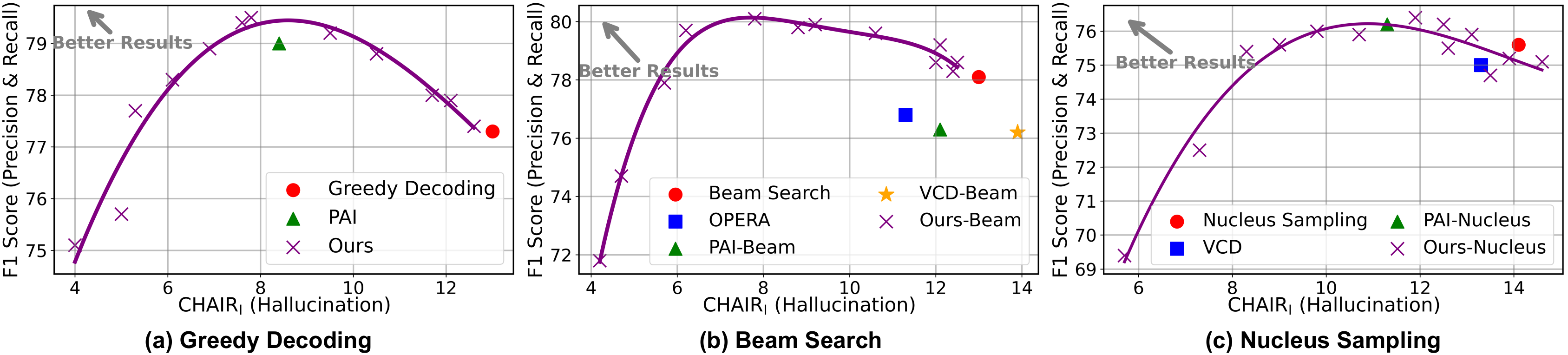}
    \vspace{-2mm}
    \caption{Ablation study on LLaVA-1.5-13B using different random seeds  ($seed$ = 48).}
    \label{fig:ablation-llava-13b-seed48}
\end{figure}

\begin{figure}[htbp]
    \centering
    \includegraphics[width=\linewidth]{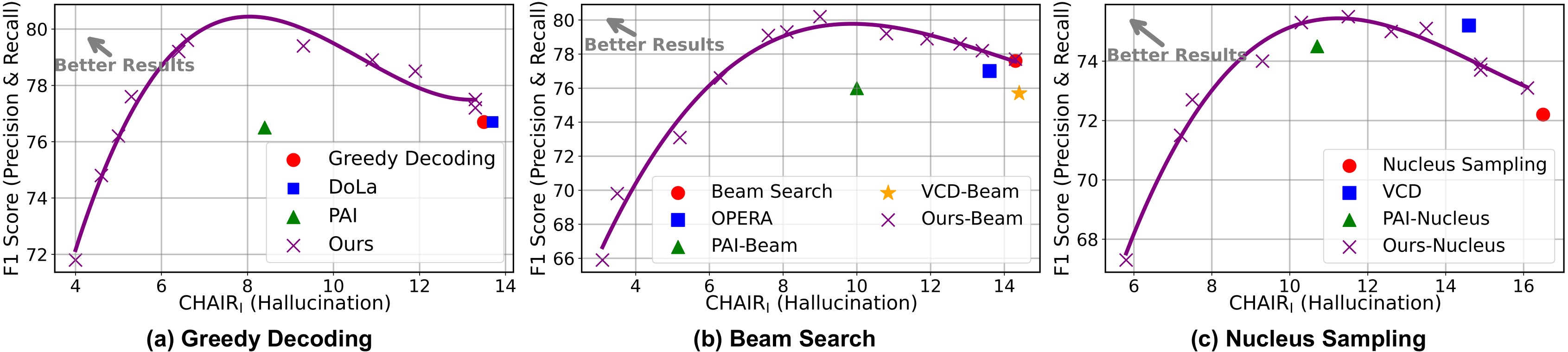}
    \vspace{-2mm}
    \caption{Ablation study on LLaVA-1.5-7B using different random seeds  ($seed$ = 104).}
    \label{fig:ablation-llava-7b-seed104}
\end{figure}

\begin{figure}[htbp]
    \centering
    \includegraphics[width=\linewidth]{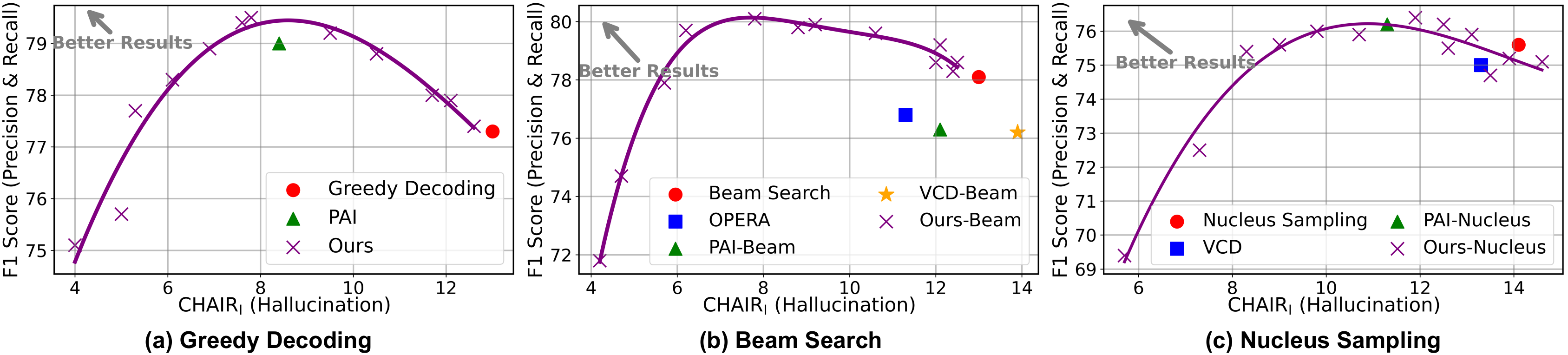}
    \vspace{-2mm}
    \caption{Ablation study on LLaVA-1.5-13B using different random seeds  ($seed$ = 104).}
    \label{fig:ablation-llava-13b-seed104}
\end{figure}

\tcj{
To mitigate the potential influence of random image subset sampling on the results of the CHAIR benchmark and to further validate the robustness of our proposed AttnReal method, we conduct additional experiments using two different random seeds, distinct from the one used in the main manuscript.
Specifically, we select LLaVA-1.5-7B and LLaVA-1.5-13B as representative models of different model scales, as most of the comparative methods have been implemented on these models.
We apply our proposed AttnReal to these models using different down-scaling factors $\alpha$ and compare the performance with existing training-free methods. 
~\cref{fig:ablation-llava-7b-seed48,fig:ablation-llava-13b-seed48} show the comparative results using the first additional random seed, and~\cref{fig:ablation-llava-7b-seed104,fig:ablation-llava-13b-seed104} demonstrate the results using the second random seed.
As can be concluded, AttnReal consistently achieves an excellent trade-off between hallucination and F1 score and can yield superior results to state-of-the-art methods on different decoding strategies using both additional random seeds.
This consistency confirms that the effectiveness of AttnReal is not contingent on specific samples, thereby strengthening the robustness of our proposed method.
}

\subsection{AMBER Benchmark}
\label{app:amber}
\tcj{
AMBER (\textbf{A}n LLM-free \textbf{M}ulti-dimensional \textbf{Be}nchma\textbf{r}k)~\cite{wang2023amber} introduces a comprehensive LLM-free framework for assessing hallucinations in MLLMs across diverse dimensions. 
The benchmark consists of 1004 samples and supports the evaluation of generative and discriminative tasks, and we focus on the generative evaluation. 
AMBER employs four metrics for evaluating hallucinations on generative tasks: \textit{CHAIR}, \textit{Cover}, \textit{Hal}, and \textit{Cog}.
Similar to CHAIR~\cite{rohrbach2018chair}, AMBER compares the mentioned objects by MLLMs and the annotated object lists and calculates \textit{CHAIR} and \textit{Cover} scores:
\begingroup
\small
\begin{equation}
\text{\textit{CHAIR}} = 1-\frac{\lvert \{\text{correct mentioned objects} \}\rvert}{\lvert \{\text{all mentioned objects} \}\rvert}, \quad
\text{\textit{Cover}} = \frac{\lvert \{\text{correct mentioned objects} \}\rvert}{\lvert \{\text{ground-truth objects} \}\rvert}
\end{equation}
\endgroup
Besides, \textit{Hal} reflects the proportion of hallucinated responses from all samples. \text{Cog} assesses how similar the hallucinations are to those in human cognition by comparing the mentioned objects by MLLMs and a set of human-curated hallucinatory target objects:
\begingroup
\small
\begin{equation}
\text{\textit{Cog}} = \frac{\lvert \{\text{hallucinated target objects} \}\rvert}{\lvert \{\text{all mentioned objects} \}\rvert}
\end{equation}
\endgroup
We evaluate our proposed AttnReal on the AMBER benchmark and various open-source MLLMs and decoding strategies. 
As can be seen in~\cref{tab: amber-5model}, our proposed AttnReal achieves the best hallucination performance (including CHAIR, Hal, and Cog metrics) across most model and decoding strategy settings, which demonstrates the effectiveness and wide applicability of AttnReal.
Meanwhile, AttnReal yields comparable or superior Cover scores compared to baselines on most models, indicating that AttnReal can improve the overall performance of MLLMs beyond response faithfulness.
}

\begin{table*}[htbp]
\caption{
    AMBER generative evaluation results on five open-source MLLMs. For CHAIR, Hal and Cog, lower values mean better performance. For Cover, high values mean better performance.
    The best results are \textbf{bolded}, and the suboptimal results are \underline{underlined}.
    ``-'' means the corresponding method is not implemented in this model.
    }
\label{tab: amber-5model}
\vspace{-2.5mm}
\resizebox{\linewidth}{!}{%
\setlength{\tabcolsep}{1.2mm}
\begin{tabular}{lcccccccccccccc}
\toprule
\multirow{2}{*}{Methods} & \multirow{2}{*}{Decoding Strategy} &
\multicolumn{4}{c}{LLaVA-1.5-7B} & \multicolumn{4}{c}{LLaVA-1.5-13B} & \multicolumn{4}{c}{MiniGPT-4} \\ 
\cmidrule(lr){3-6} \cmidrule(lr){7-10} \cmidrule(lr){11-14}
 &  & 
  CHAIR$\downarrow$ &  Cover$\uparrow$ &  Hal$\downarrow$ &  Cog$\downarrow$ &
  CHAIR$\downarrow$ &  Cover$\uparrow$ &  Hal$\downarrow$ &  Cog$\downarrow$ &
  CHAIR$\downarrow$ &  Cover$\uparrow$ &  Hal$\downarrow$ &  Cog$\downarrow$ \\
\cmidrule(lr){1-14}

Greedy           & \multirow{4}{*}{Greedy Decoding} & 6.1 & \underline{51.1} & 28.9 & 3.1 & 6.2 & 51.0 & 28.9 & 3.0 & 15.1 & \textbf{62.6} & 62.8 & 10.3  \\
DoLa~\cite{chuang2023dola}             &                                   & 6.3 & \underline{51.1} & 29.2 & 3.1 & - & - & - & - & 16.0 & \textbf{62.6} & 63.4 & 11.0  \\
PAI~\cite{liu2024pai}              &                                   & \underline{4.5} & 47.2 & \underline{21.0} & \textbf{1.4} & \underline{4.8} & \underline{51.5} & \underline{26.3} & \underline{1.6} & \underline{10.7} & \underline{59.3} & \underline{43.0} & \underline{5.6}  \\
\textbf{Ours}  &         & \textbf{3.2} & \textbf{51.2} & \textbf{19.2} & \underline{1.5} & \textbf{3.2} & \textbf{51.9} & \textbf{18.9} & \textbf{1.5} & \textbf{10.3} & 56.7 & \textbf{41.7} & \textbf{5.2}     \\

\cmidrule(lr){1-14}
Beam Search      & \multirow{5}{*}{Beam Search}      & 7.4 & \underline{49.4} & 30.9 & 3.6 & 7.2 & \underline{49.5} & 30.8 & 3.6 & 14.1 & \underline{61.0} & 60.1 & 9.1  \\
OPERA~\cite{huang2024opera}            &                                   & \underline{6.4} & 48.9 & 27.7 & 3.0 & \underline{6.1} & 48.1 & \underline{24.8} & \underline{2.7} & 12.3 & 57.7 & 48.1 & 6.4  \\
VCD-Beam~\cite{leng2024vcd}         &                                   & 7.5 & \textbf{51.2} & 33.5 & 3.3 & 7.5 & \textbf{50.3} & 32.6 & 3.0 & 15.4 & \textbf{61.5} & 61.8 & 9.4  \\
PAI-Beam~\cite{liu2024pai}         &                                   & 7.2 & 45.3 & \underline{27.6} & \underline{2.4} & 6.2 & 47.6 & 26.9 & \textbf{1.8} & \underline{10.1} & 55.2 & \underline{36.2} & \underline{4.7}  \\
\textbf{Ours}  &         & \textbf{4.1} & 48.6 & \textbf{21.6} & \textbf{1.8} & \textbf{4.2} & 48.0 & \textbf{20.8} & \textbf{1.8} & \textbf{7.8} & 55.7 & \textbf{34.6} & \textbf{3.9}    \\

\cmidrule(lr){1-14}
Nucleus Sampling & \multirow{4}{*}{Nucleus Sampling} & 9.0 & \textbf{50.6} & 40.4 & 3.6 & 8.6 & 49.6 & 38.4 & 3.6 & 15.5 & \underline{60.2} & 63.5 & 9.4  \\
VCD~\cite{leng2024vcd}              &                                   & 7.3 & \underline{50.2} & 33.8 & \underline{3.1} & 7.3 & \textbf{50.7} & 32.6 & 3.3 & 15.3 & \textbf{60.9} & 61.7 & 9.6  \\
PAI-Nucleus~\cite{liu2024pai}      &                                   & \underline{6.6} & 49.2 & \textbf{31.1} & \textbf{2.2} & \underline{6.6} & \underline{50.6} & \underline{31.3} & \underline{2.2} & \underline{11.3} & 57.4 & \underline{43.8} & \underline{5.8}  \\
\textbf{Ours}  &         & \textbf{6.4} & 48.2 & \underline{31.9} & \textbf{2.2} & \textbf{5.4} & 50.3 & \textbf{29.0} & \textbf{1.9} & \textbf{10.4} & 54.7 & \textbf{42.0} & \textbf{4.4}    \\

\bottomrule
\end{tabular}
}

\vspace{1mm}

\resizebox{\linewidth}{!}{%
\setlength{\tabcolsep}{2.5mm}
\begin{tabular}{lccccccccc}
\toprule
\multirow{2}{*}{Methods} & \multirow{2}{*}{Decoding Strategy} &
\multicolumn{4}{c}{Shikra} & \multicolumn{4}{c}{mPLUG-Owl2} \\ 
\cmidrule(lr){3-6} \cmidrule(lr){7-10}
 &  & 
  CHAIR$\downarrow$ &  Cover$\uparrow$ &  Hal$\downarrow$ &  Cog$\downarrow$ &
  CHAIR$\downarrow$ &  Cover$\uparrow$ &  Hal$\downarrow$ &  Cog$\downarrow$ \\
\cmidrule(lr){1-10}

Greedy           & \multirow{4}{*}{Greedy Decoding}  & 11.0 & \textbf{50.9} & 48.2 & 5.2 & \underline{9.9} & \underline{51.5} & \underline{40.3} & \underline{5.0} \\
DoLa~\cite{chuang2023dola}             &                                   & - & - & - & - & 10.3 & 51.4 & 42.4 & 5.2  \\
PAI~\cite{liu2024pai}              &                                   & \underline{6.5} & 49.0 & \underline{32.4} & \underline{2.3} & - & - & - & -    \\
\textbf{Ours}  &         & \textbf{5.2} & \underline{49.6} & \textbf{27.9} & \textbf{2.1} & \textbf{5.4} & \textbf{52.0} & \textbf{26.2} & \textbf{2.2}    \\

\cmidrule(lr){1-10}
Beam Search      & \multirow{5}{*}{Beam Search}      & 10.7 & \textbf{50.9} & 47.7 & 5.8 & 11.5 & 49.3 & 42.3 & 5.3    \\
OPERA~\cite{huang2024opera}            &                                   & 9.5 & \underline{50.1} & 41.2 & 4.5 & \underline{10.1} & 49.0 & \underline{36.9} & \underline{4.1}    \\
VCD-Beam~\cite{leng2024vcd}         &                                   & - & - & - & - & 11.5 & \textbf{51.2} & 45.4 & 5.0   \\
PAI-Beam~\cite{liu2024pai}         &                                   & \underline{7.2} & 47.5 & \underline{30.8} & \underline{2.8} & - & - & - & -   \\
\textbf{Ours}  &         & \textbf{4.8} & 48.0 & \textbf{25.0} & \textbf{1.9} & \textbf{6.8} & \underline{49.8} & \textbf{26.5} & \textbf{2.6}    \\

\cmidrule(lr){1-10}
Nucleus Sampling & \multirow{4}{*}{Nucleus Sampling} & 11.7 & \textbf{51.8} & 53.5 & 5.7 & 13.0 & 50.3 & 50.0 & 5.4  \\
VCD~\cite{leng2024vcd}               &                                   & - & - & - & - & \underline{11.5} & \underline{51.3} & \underline{45.7} & \underline{4.9}   \\
PAI-Nucleus~\cite{liu2024pai}       &                                   & \underline{7.0} & 49.1 & \underline{33.0} & \textbf{2.4} & - & - & - & -  \\
\textbf{Ours}  &         & \textbf{5.4} & \underline{50.5} & \textbf{29.8} & \underline{2.5} & \textbf{9.4} & \textbf{51.8} & \textbf{40.7} & \textbf{3.2}    \\

\bottomrule
\end{tabular}%
}
\vspace{-4mm}
\end{table*}

\subsection{MMHal-Bench}
\label{app:mmhal}
\begin{figure*}[htbp]
    \centering
    \includegraphics[width=\linewidth]{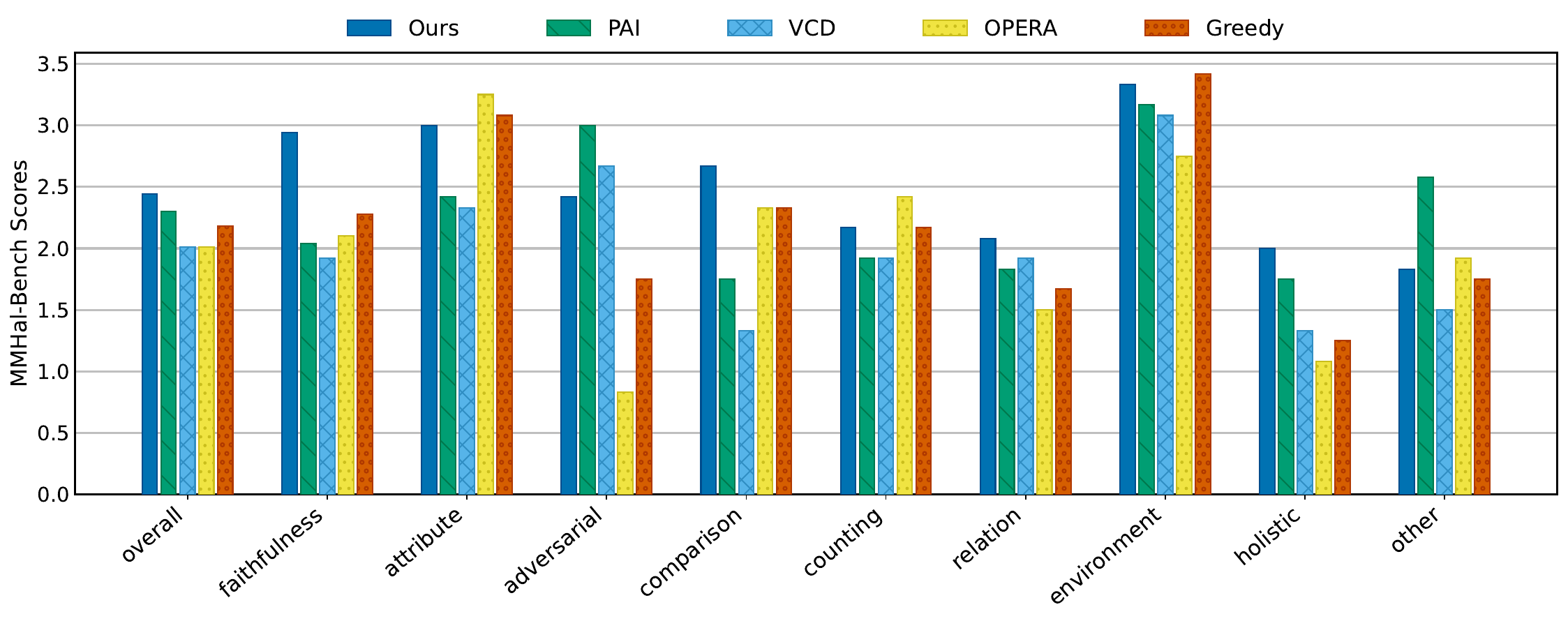}
    \caption{Comparative results on the MMHal-Bench and LLaVA-1.5-13B. To all metrics, higher scores mean better performance.}
    \label{fig:mmhal-13b}
    \vspace{6mm}
\end{figure*}

\begin{figure*}[htbp]
    \centering
    \includegraphics[width=\linewidth]{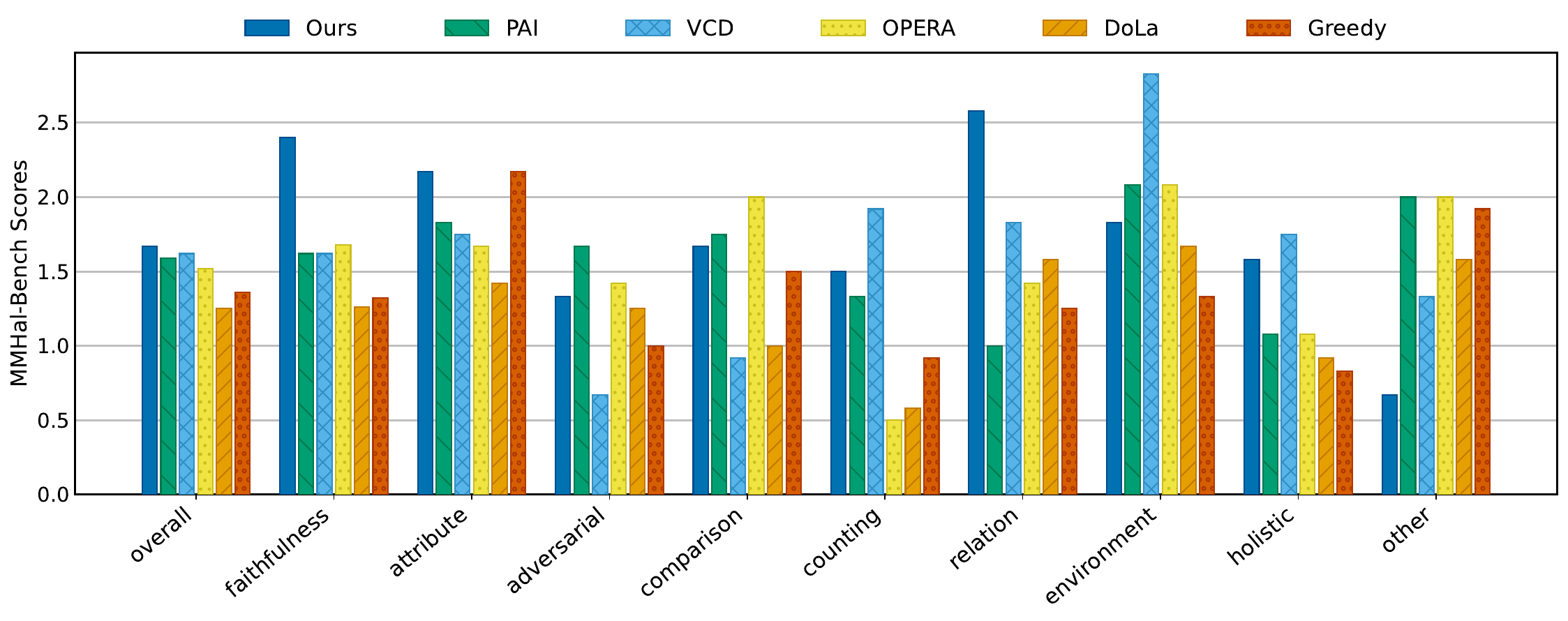}
    \caption{Comparative results on the MMHal-Bench and MiniGPT-4. To all metrics, higher scores mean better performance.}
    \label{fig:mmhal-minigpt}
    \vspace{6mm}
\end{figure*}

\begin{figure*}[htbp]
    \centering
    \includegraphics[width=\linewidth]{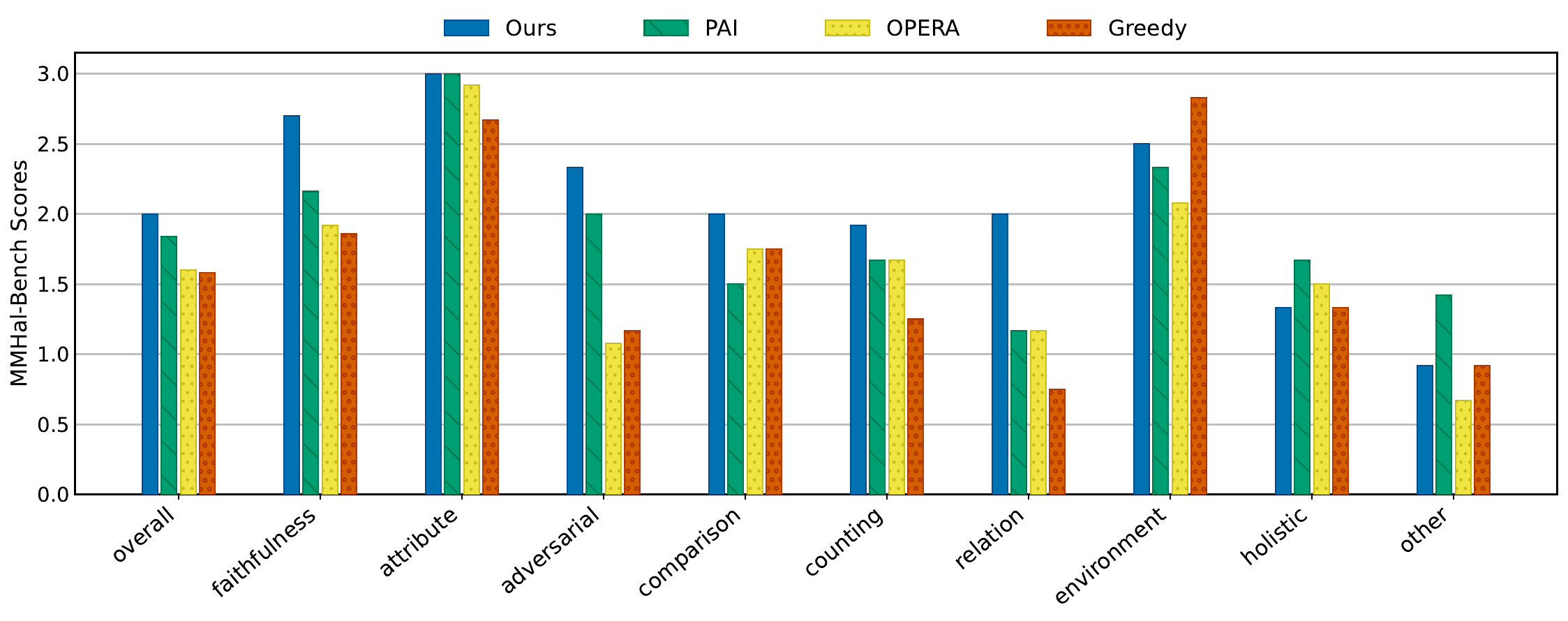}
    \caption{Comparative results on the MMHal-Bench and Shikra. To all metrics, higher scores mean better performance.}
    \label{fig:mmhal-shikra}
    \vspace{6mm}
\end{figure*}

\begin{figure}[htbp]
    \centering
    \includegraphics[width=\linewidth]{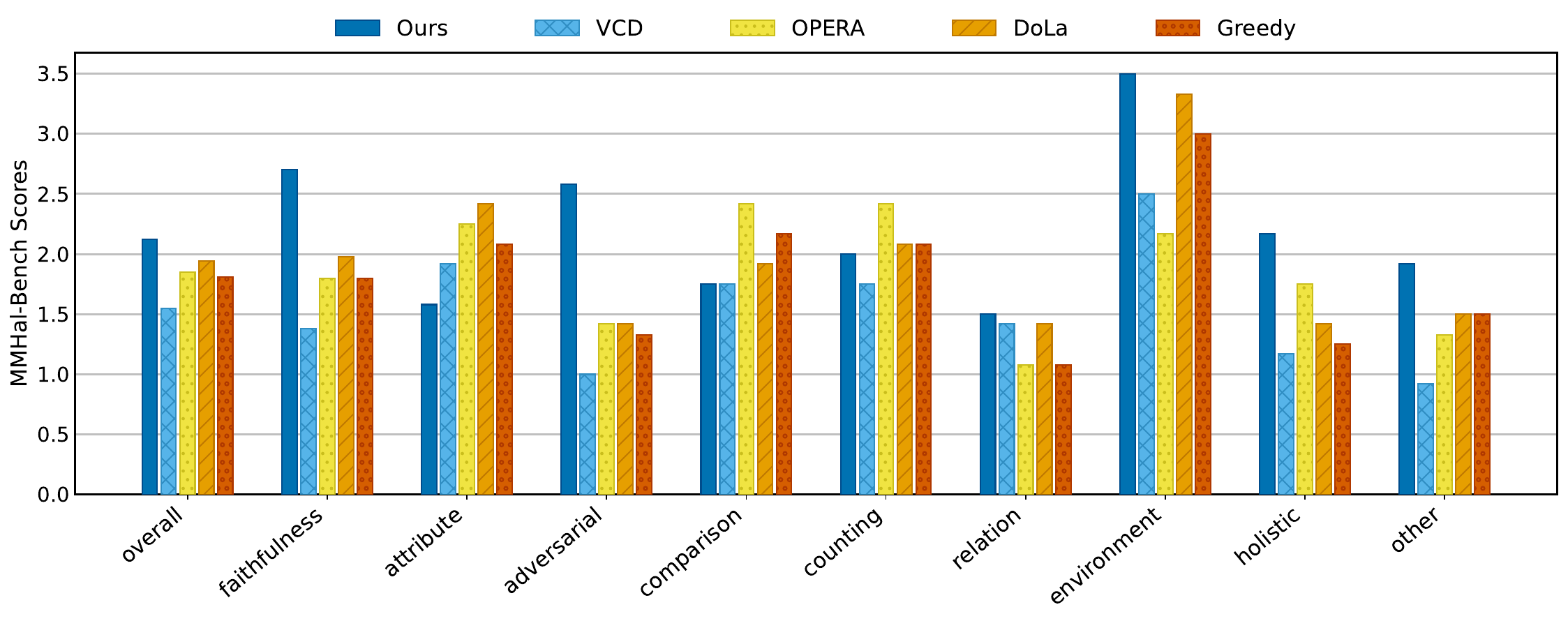}
    \caption{Comparative results on the MMHal-Bench and mPLUG-Owl2. To all metrics, higher scores mean better performance.}
    \label{fig:mmhal-mplug}
    \vspace{6mm}
\end{figure}

In this section, the results of more MLLMs on MMHal-Bench are demonstrated in~\cref{fig:mmhal-13b,fig:mmhal-minigpt,fig:mmhal-shikra,fig:mmhal-mplug}.
On all models, our proposed AttnReal method achieves the highest overall scores and substantially improves response faithfulness, demonstrating the comprehensive capability of our AttnReal method on various hallucination types.

\newpage

\section{Qualitative Study}
\label{sec:qualitative study}
\begin{figure*}[htbp]
    \centering
    \includegraphics[width=0.76\linewidth]{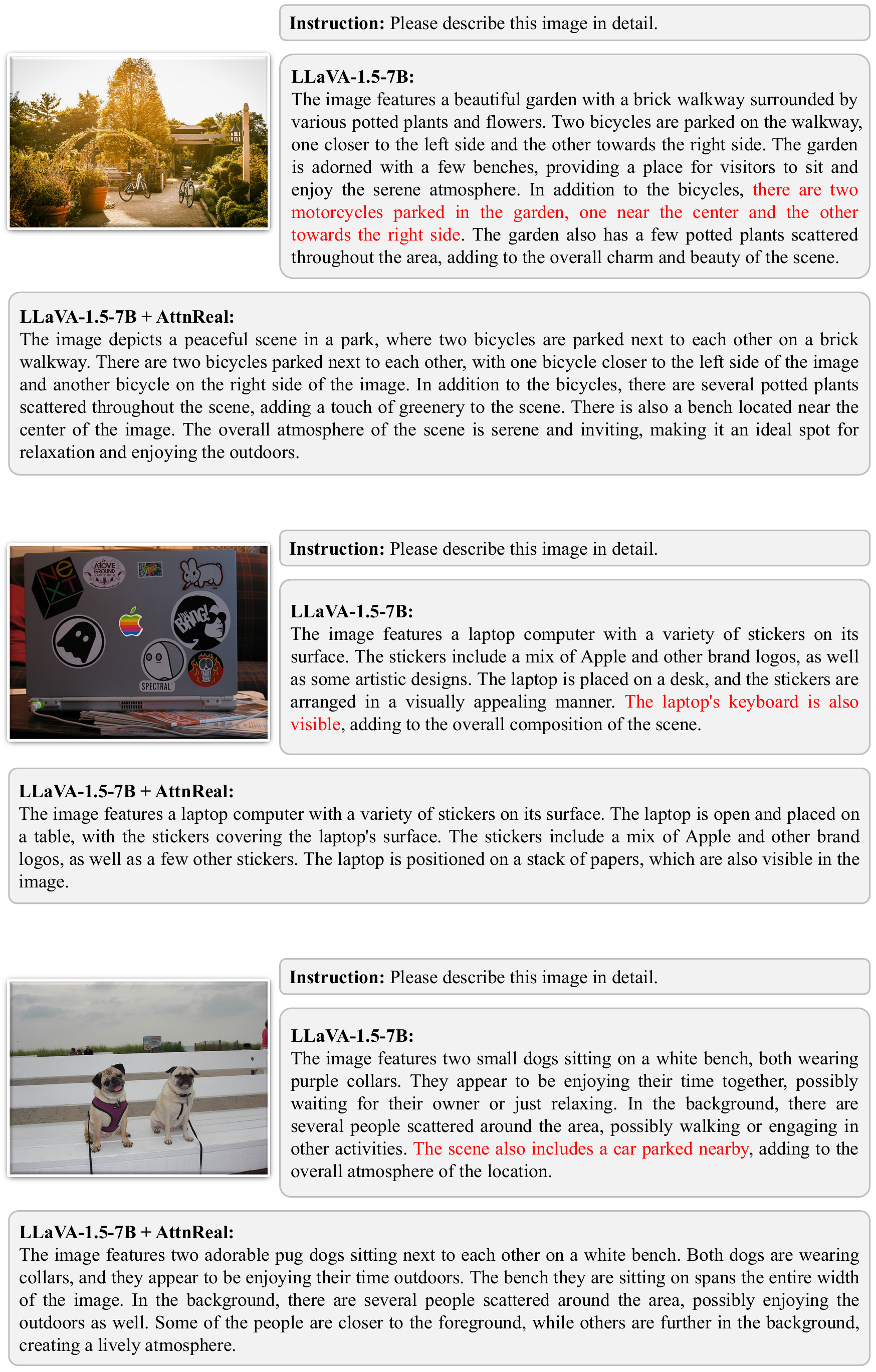}
    \caption{
    Qualitative results of the hallucination mitigation effect of our proposed AttnReal on the LLaVA-1.5-7B model.
    Hallucinations in the responses are highlighted in \red{red}.
    }
    \label{fig:case-llava-7b}
\end{figure*}

\begin{figure*}[htbp]
    \centering
    \includegraphics[width=0.8\linewidth]{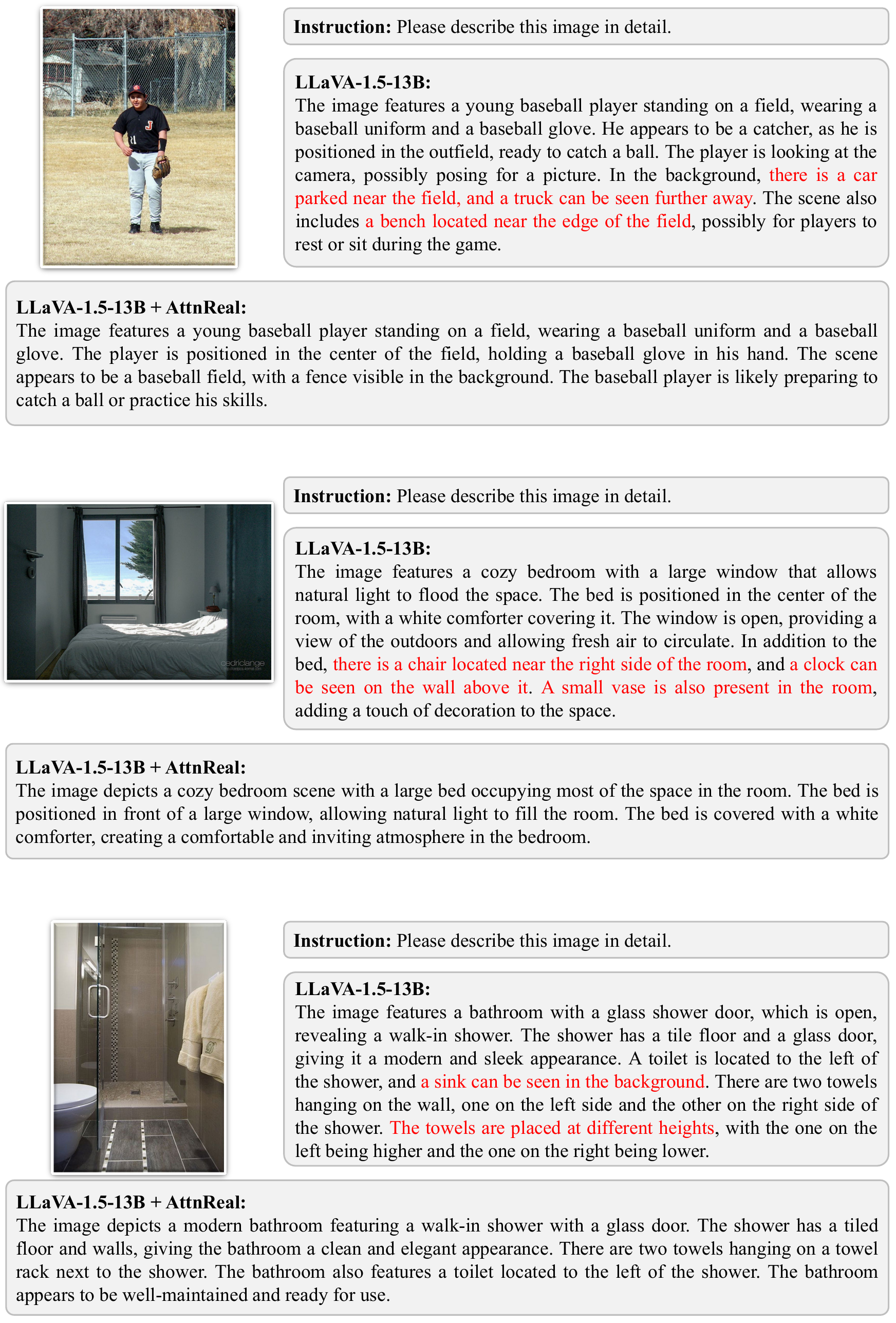}
    \caption{
    Qualitative results of the hallucination mitigation effect of our proposed AttnReal on the LLaVA-1.5-13B model.
    Hallucinations in the responses are highlighted in \red{red}.
    }
    \label{fig:case-llava-13b}
\end{figure*}

\begin{figure*}[htbp]
    \centering
    \includegraphics[width=0.8\linewidth]{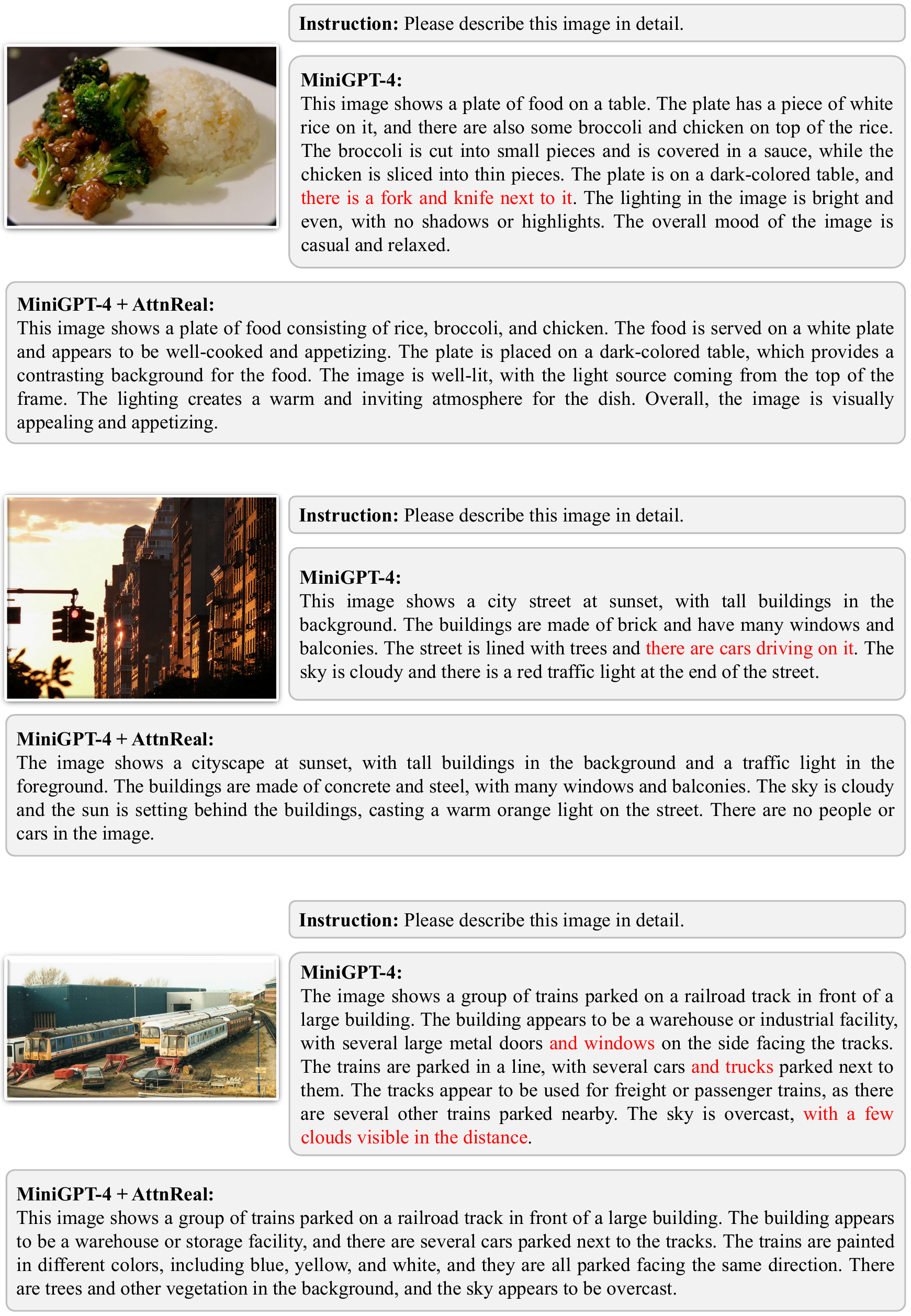}
    \caption{
    Qualitative results of the hallucination mitigation effect of our proposed AttnReal on the MiniGPT-4 model.
    Hallucinations in the responses are highlighted in \red{red}.
    }
    \label{fig:case-minigpt}
\end{figure*}

\begin{figure*}[htbp]
    \centering
    \includegraphics[width=0.78\linewidth]{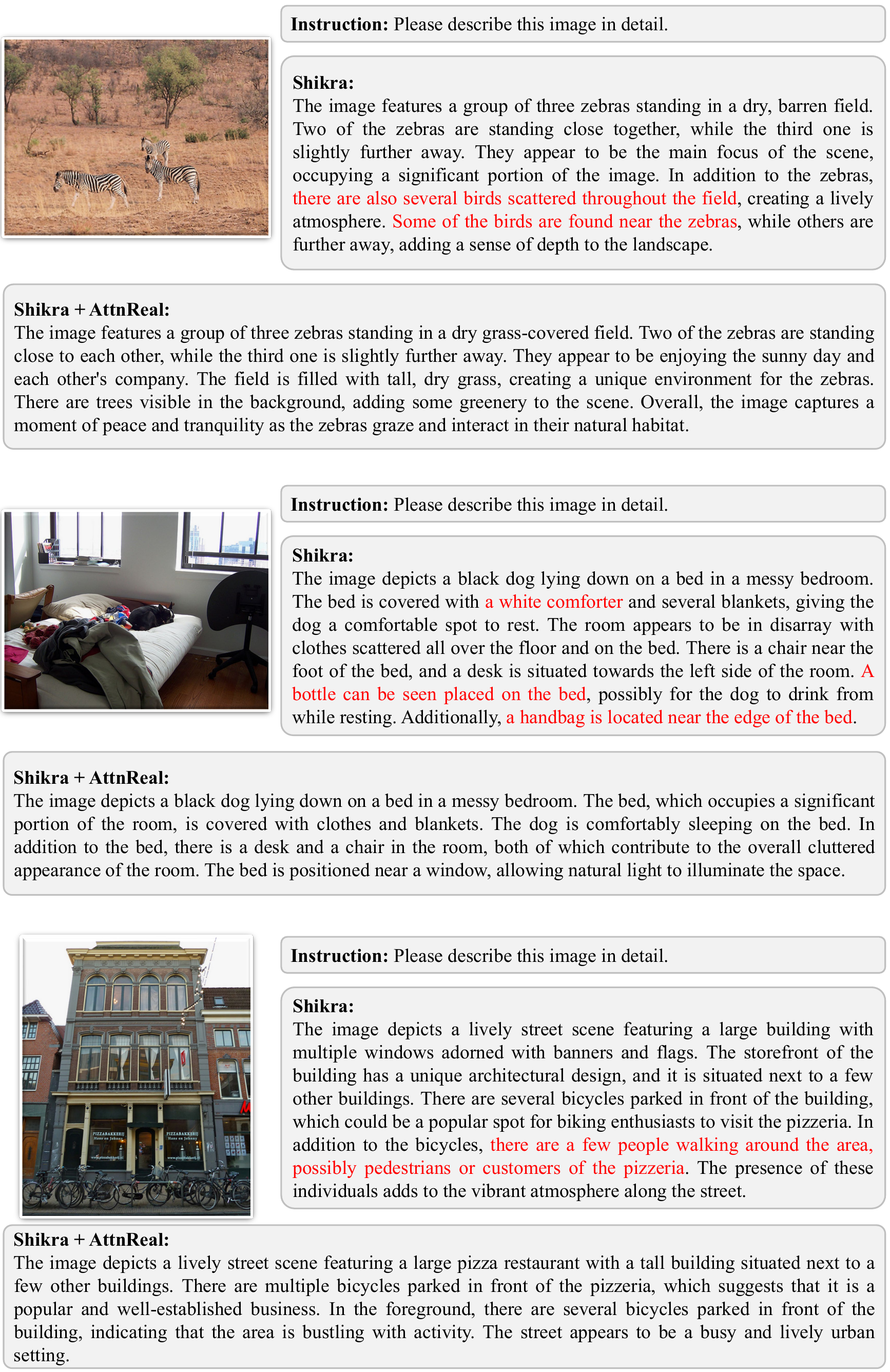}
    \caption{
    Qualitative results of the hallucination mitigation effect of our proposed AttnReal on the Shikra model.
    Hallucinations in the responses are highlighted in \red{red}.
    }
    \label{fig:case-shikra}
\end{figure*}

\begin{figure*}[htbp]
    \centering
    \includegraphics[width=0.76\linewidth]{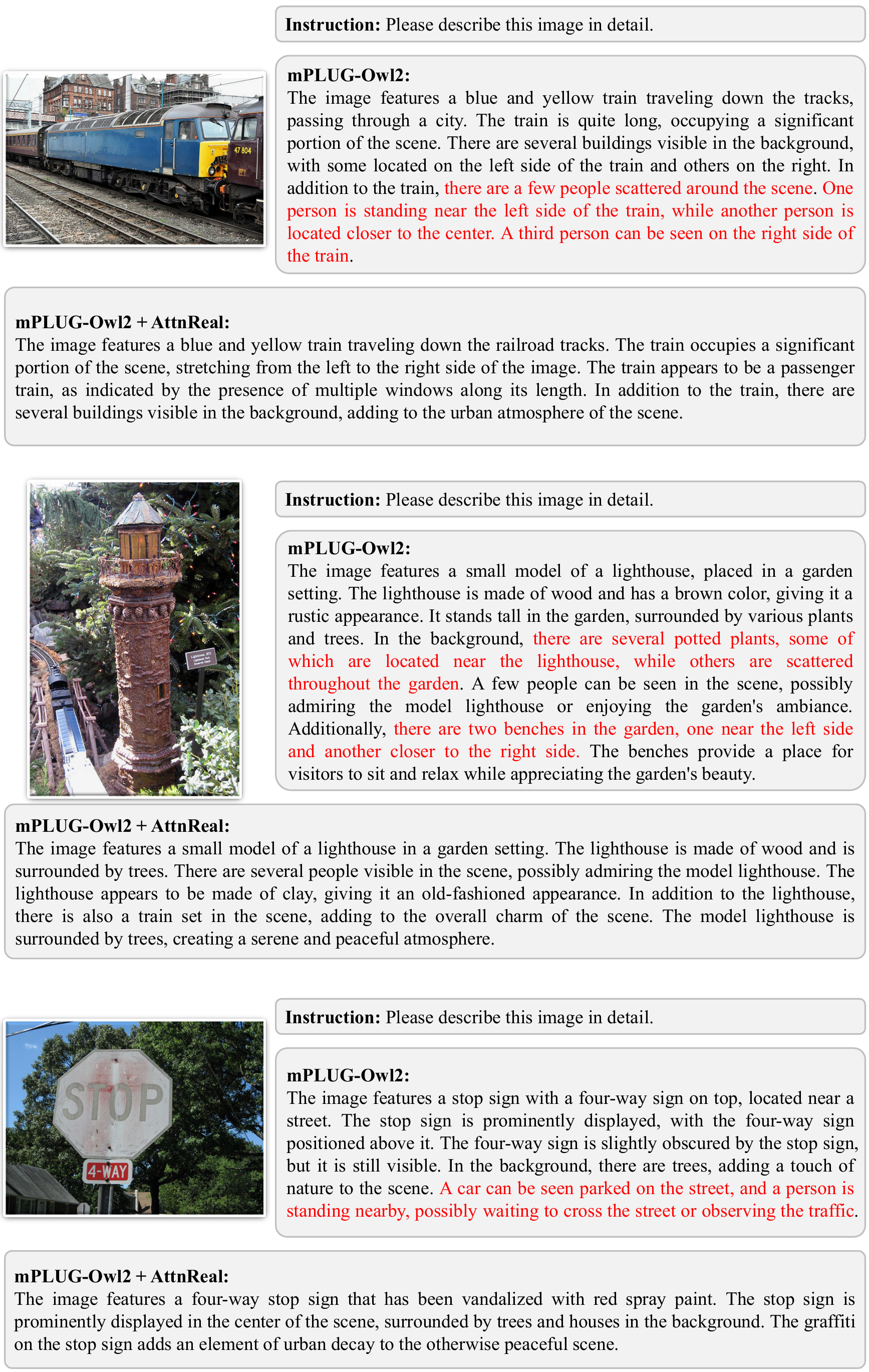}
    \caption{
    Qualitative results of the hallucination mitigation effect of our proposed AttnReal on the mPLUG-Owl2 model.
    Hallucinations in the responses are highlighted in \red{red}.
    }
    \label{fig:case-mplug}
\end{figure*}

\begin{figure*}[htbp]
    \centering
    \includegraphics[width=0.76\linewidth]{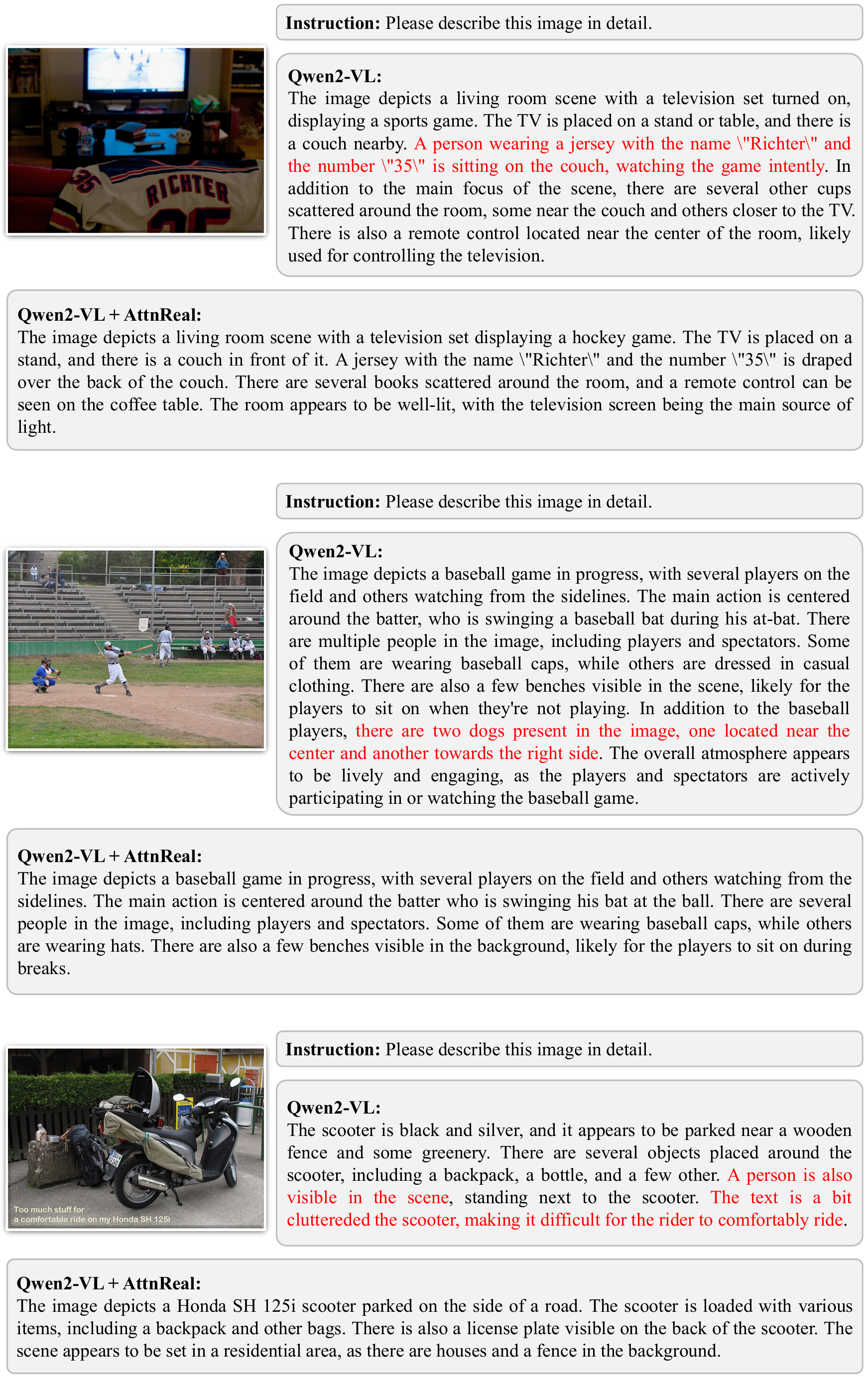}
    \caption{
    Qualitative results of the hallucination mitigation effect of our proposed AttnReal on the Qwen2-VL model.
    Hallucinations in the responses are highlighted in \red{red}.
    }
    \label{fig:case-qwen}
\end{figure*}

In this section, we show the qualitative results of AttnReal eliminating the hallucinations of multiple MLLMs.
Specifically, the hallucinated responses of the MLLMs and the corrected responses when applying AttnReal are provided, as well as the input images. 
~\cref{fig:case-llava-7b,fig:case-llava-13b,fig:case-minigpt,fig:case-shikra,fig:case-mplug,fig:case-qwen} correspond to LaVA-1.5-7B~\cite{liu2024visualllava}, LLaVA-1.5-13B~\cite{liu2024visualllava}, MiniGPT-4~\cite{zhu2023minigpt}, Shikra~\cite{chen2023shikra}, mPLUG-Owl2~\cite{ye2024mplug}, and Qwen2-VL~\cite{wang2024qwen2}, respectively, with three cases for each model. Hallucinations in the responses are highlighted in red.
From the results, we can conclude that the responses of MLLMs are more faithful and more focused on the image contents using AttnReal.

\newpage

\section{GPT-assisted Evaluation Prompt}
\label{app:gpt prompt}
We provide the prompt used for GPT-assisted evaluation in~\cref{tab:gpt_prompt}, which is constructed based on the prompt used by PAI~\cite{liu2024pai}. Specifically, we follow MMHal-Bench~\cite{sun2023aligning_train_mmhal} to specify the criterion for each rating in the prompt to conduct a fairer evaluation of multiple MLLMs and methods.

\begin{table*}[htb]
\caption{Prompt for GPT-4o assisted evaluation.}
\label{tab:gpt_prompt}
\vspace{-2mm}
\begin{tabular}{p{0.97\textwidth}}
\hline
\textbf{GPT-4o Assisted Evaluation} \\
\hline
 You are required to score the performance of an AI assistant in describing a given image. Please pay extra attention to hallucination, which refers to the part of descriptions that are inconsistent with the image content, such as claiming the existence of something not present in the image or describing incorrectly in terms of the counts, positions, or colors of objects in the image. Please rate the response of the assistant on a scale of 1 to 5, where a higher score indicates better performance, according to the following criteria:\\
\\
\textbf{Correctness}\\
 5 - The response is entirely correct, with no hallucinations.\\
 4 - The response is mostly correct, with only minor inaccuracies that do not significantly alter the meaning.\\
 3 - The response contains some inaccuracies, but they are not critical to understanding the main content.\\
 2 - The response has multiple inaccuracies that affect the correctness.\\
 1 - The response is largely incorrect, with significant hallucinations.\\
\\
\textbf{Detailedness}\\
 5 - The response is rich in necessary details, covering all important aspects of the image without irrelevant information.\\
 4 - The response provides a good level of detail, but misses a minor aspect.\\
 3 - The response includes some details but misses multiple important aspects.\\
 2 - The response is lacking in important details and is vague.\\
 1 - The response provides minimal detail and lacks depth.\\
\\
Please output two scores for the AI assistant, indicating the ratings for Correctness and Detailedness respectively. Following the scores, please provide an explanation of your evaluation.\\
\\
 $[$Assistant Response$]$\\
 \{response\}\\
 $[$End of Assistant Response$]$\\
\\
 Output format:\\
 Correctness: $<$Score 1$>$\\
 Detailedness: $<$Score 2$>$\\
 Reason:\\
\hline
\end{tabular}
\end{table*}

\section{Algorithm of Attention Reallocation}
\label{app:algorithm table}
In this section, we provide the algorithm of our proposed Attention Reallocation (AttnReal) approach.

\begin{algorithm}[htbp]
\caption{Attention Reallocation Steps}
\begin{algorithmic}[1]

\Statex \hspace{-4.5mm} \textbf{Input:} Post-softmax attention map of all token types $[A_s, A_v, A_i, A_o]$

\State \textbf{Identify Sinks:} Construct the attention sink set $S_{\text{sink}}$ as~\cref{eq:identify_sinks}
\State \textbf{Recycle Attention:} Suppress $A_o$ for tokens in $S_{\text{sink}}$  as~\cref{eq:recycle_attn}
\State \textbf{Reallocate Attention:} Strengthen $A_v$ as~\cref{eq:allocate_attn}

\Statex \hspace{-4.5mm} \textbf{Output:} Updated attention map $[A_s, A'_v, A_i, A'_o]$
\end{algorithmic}
\end{algorithm}

\end{document}